\pdfoutput=1

\documentclass[11pt]{article}

\usepackage[final]{acl}

\usepackage{times}
\usepackage{latexsym}
\usepackage{array}
\usepackage{graphicx}

\usepackage{enumitem}

\usepackage[most]{tcolorbox}

\usepackage{mathtools}
\usepackage{enumitem}
\usepackage{colortbl}
\usepackage{diagbox}
\usepackage{wrapfig}
\usepackage{pst-node} 
\usepackage{makecell}

\usepackage[T1]{fontenc}

\usepackage[utf8]{inputenc}

\usepackage{microtype}

\usepackage{inconsolata}

\usepackage{tabularx}
\usepackage{multirow}
\usepackage{booktabs}
\usepackage[normalem]{ulem}
\usepackage{xspace}
\usepackage{authblk}
\renewcommand{\footnotesize}{\small}
\renewcommand{\arraystretch}{0.9}
\title{HOLMES: Hyper-Relational Knowledge Graphs for Multi-hop Question Answering using LLMs}

\author[1]{\textbf{Pranoy Panda} \thanks{Corresponding author : \texttt{pranoy.panda@fujitsu.com}}}
\author[1]{\textbf{Ankush Agarwal}}
\author[1]{\textbf{Chaitanya Devaguptapu}}
\author[1]{\\ \textbf{Manohar Kaul}}
\author[1, 2]{\textbf{Prathosh A P}}

\affil[1]{Fujitsu Research India}
\affil[2]{Indian Institute of Science, Bengaluru}
\affil[ ]{\texttt{\{pranoy.panda, ankush.agarwal, manohar.kaul\}@fujitsu.com}}
\affil[ ]{\texttt{email@chaitanya.one}, \texttt{prathoshap@gmail.com}
}

\newcommand{\name}{HOLMES\xspace}
\newcommand{\updatedText}{black}

\begin{document}
\maketitle
\begin{abstract}
Given unstructured text, Large Language Models (LLMs) are adept at answering simple (single-hop) questions. However, as the complexity of the questions increase, the performance of LLMs degrade. We believe this is due to the overhead associated with understanding the complex question followed by filtering and aggregating unstructured information in the raw text. Recent methods try to reduce this burden by integrating structured knowledge triples into the raw text, aiming to provide a structured overview that simplifies information processing. However, this simplistic approach is query-agnostic and the extracted facts are ambiguous as they lack context. To address these drawbacks and to enable LLMs to answer complex (multi-hop) questions with ease, we propose to use a knowledge graph (KG) that is \emph{context-aware} and is distilled to contain query-relevant information. 
\textcolor{\updatedText}{The use of our compressed distilled KG as input to the LLM results in our method utilizing up to $67\%$ fewer tokens to represent the query relevant information present in the supporting documents, compared to the state-of-the-art (SoTA) method.}
Our experiments show consistent improvements over the SoTA across several metrics (EM, F1, BERTScore, and Human Eval) on two popular benchmark datasets (HotpotQA and MuSiQue). 

\end{abstract}

\section{Introduction}
Multi-Hop Question Answering (MHQA) is a field that presents unique challenges in the realm of natural language processing. 
To illustrate the challenges of MHQA, consider extracting information from data arising during board meetings.
While current technologies (such as LLMs) are proficient at addressing simple (single-hop) questions, such as "\textit{How many board meetings were held in the last twelve months?}", they falter when confronted with complex (multi-hop) questions. An example of a multi-hop question is "\textit{For the board meeting with the most divided votes in the last twelve months, what was the agenda, who voted against it, and by what margin did it pass or fail?}". Answering this question requires a series of interconnected steps: first, identifying the meeting with the most divided votes; next, determining the main agenda of that meeting; then, listing the members who voted against it; and finally, calculating the margin by which it was approved or rejected. Each of these steps, or "hops", demands not only the retrieval of additional information but also a nuanced understanding of the context and the relationships between various entities. This complexity underscores the challenge of MHQA, where the goal is to navigate through layers of information to arrive at the answer. Our work is aimed at tackling these multifaceted questions.

\begin{figure}[t]
  \centering
  \includegraphics[width=0.9\columnwidth]{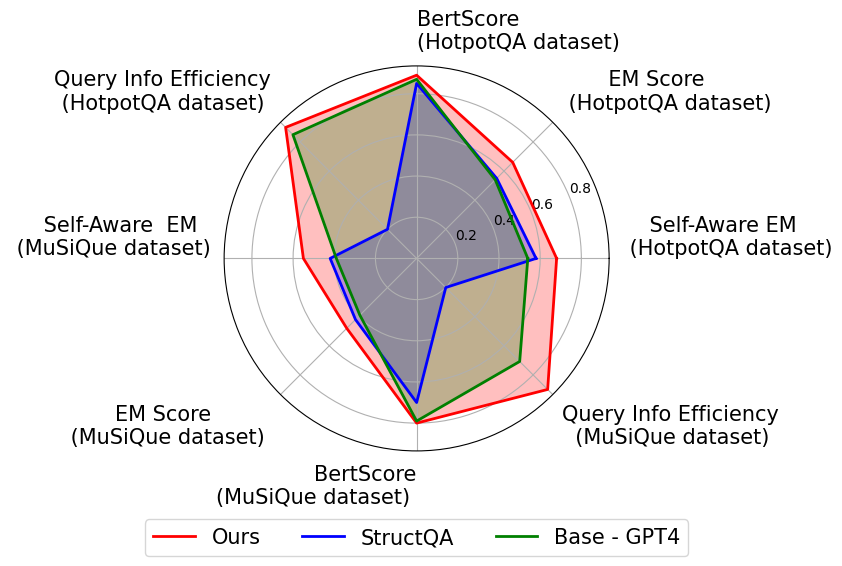}
  \caption{\textbf{Multi-Dimensional Improvements}: Our method (with GPT-4 as reader LLM) achieves SoTA results on several datasets and multiple Multi-hop QA metrics. \textit{EM}: Exact-Match with the gold answer, \textit{Self-Aware EM}: Confidence-aware EM, \textcolor{\updatedText}{\textit{BertScore} \cite{zhang2019bertscore}: Semantic similarity between predicted and gold answer; \textit{Query Info Efficiency}: Efficiency of representing query-relevant information in the supporting documents - inversely proportional to the input token count for the reader LLM.}}
  \label{fig:radar_plot}
  \vspace{-0.4cm}
\end{figure}

\begin{figure*}
  \centering
    \includegraphics[width=0.9\textwidth]{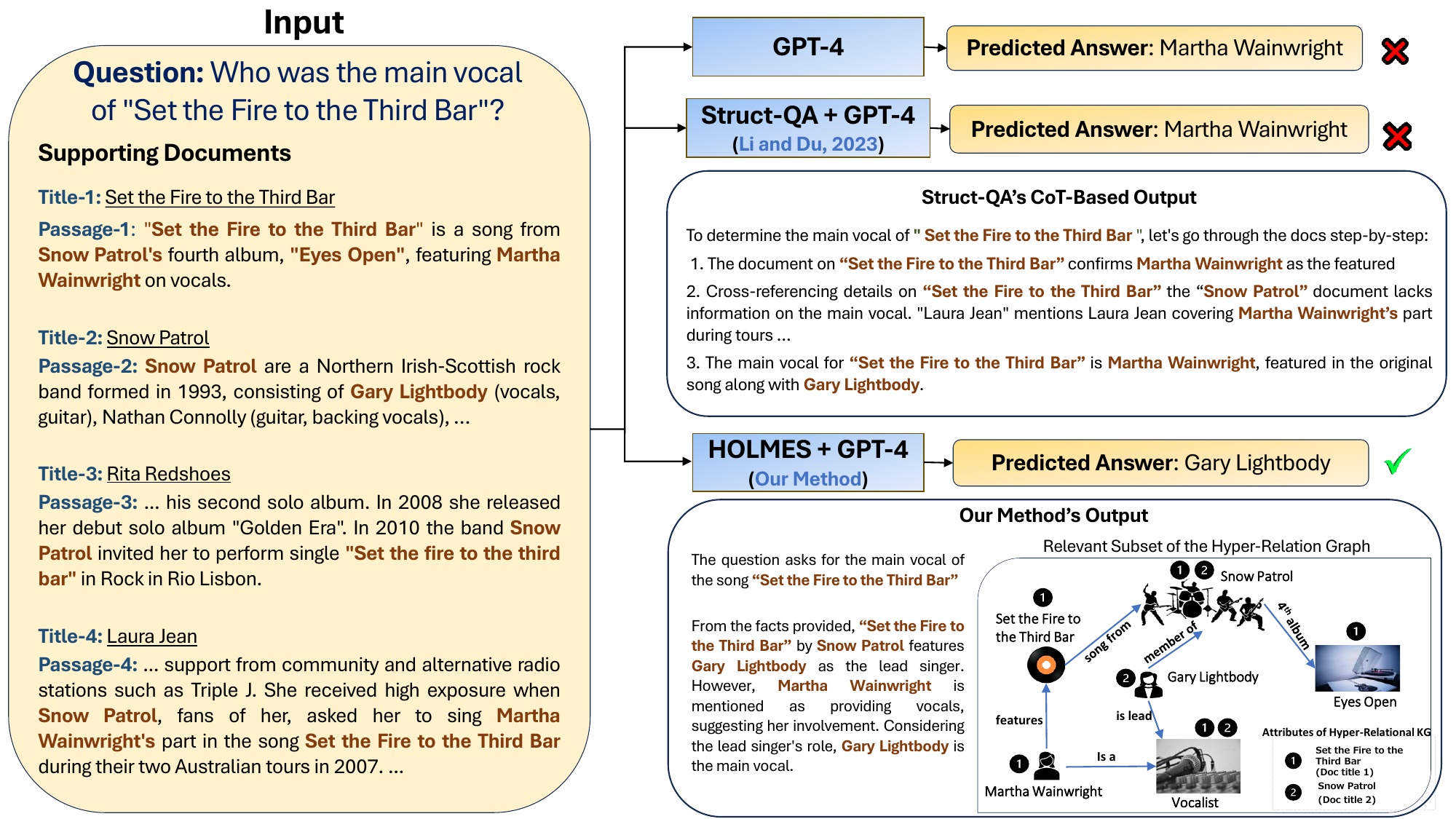}
  \caption{\textbf{Multi-Hop QA Case Study}: This figure illustrates a bridge-type multi-hop question from HotpotQA dataset for comparing our method with the baselines. It highlights our method's accurate identification of `\textit{Snow Patrol}' as the crucial bridge entity and subsequently finding the lead vocalist, a feat not achieved by baselines.}
  \label{fig:intro}
\end{figure*}



\noindent LLMs have emerged as a powerful set of tools for various NLP tasks~\cite{zhang2023sentiment,wadhwa2023revisiting,robinson2022leveraging,xu2023large}. However, despite their prowess, LLMs oftentimes are found lacking, when confronted with complex, multi-hop questions (see Figure \ref{fig:radar_plot} for an example).  We hypothesize that the degradation in performance is due to the complexity involved in \textit{filtering and contextual aggregation of information present in the unstructured text}.  


\noindent To address this challenge with unstructured text, recent methods, such as StructQA \cite{li2023leveraging}, extract structured knowledge in the form of KG triples from raw text and couple it with Chain-of-Thought (CoT) reasoning \cite{wei2022chain}. 
However, the extracted KG triples are not dependent on the query and lack the context under which these facts are valid. To understand the ambiguity due to the lack of context, consider this KG triple: \{\textit{subject: "Apple", relation: "prices rose", object: "10\%"}\}. Without additional context, it is difficult to determine whether the entity "\textit{Apple}" refers to the fruit or the company. Moreover, StructQA provides both the extracted KG triples and the raw text as input to the LLM, leading to significantly longer prompts (see Table \ref{tab:token_count}) and information redundancy. 

\noindent Our method, 
\name\footnote{Named after Sherlock Holmes for its ability to deduce query relevant information from unstructured text}
, addresses these challenges by creating a query-focused context-aware KG and using it as the sole input for the LLM, \textit{i.e.}, without inputting the raw text. Specifically, we (i) synthesize a hyper-relational KG from unstructured text that captures both facts, and the context under which these facts are valid, and (ii) prune the hyper-relational KG using a knowledge schema that encodes the type of information necessary to answer the query.
These steps collectively furnish the LLM with a curated set of relevant facts. As an example, we provide a case study on the HotpotQA dataset in Figure \ref{fig:intro}.

\noindent To rigorously evaluate our approach, we use two challenging multi-hop QA datasets, HotpotQA \cite{yang2018hotpotqa} and MuSiQue \cite{trivedi2022musique}, and experiment with three SoTA reader LLMs for QA: GPT-3.5, GPT-4, and Gemini-Pro. We compare with the baselines across $6$ metrics: EM, F1, Precision, Recall, Human-Eval, BERTScore, and achieve significant gains.
Our contributions are as follows: 

\begin{itemize}[nosep]
    \item A new multi-hop QA approach that transforms unstructured text into a hyper-relational KG using a query-derived schema, serving as an input to the LLM.
    \item A significant improvement over the SoTA multi-hop QA method \cite{li2023leveraging}:   18.7\% and 20\% in EM scores on the Hotpot dataset, and 26\% and 14.3\% on the MuSiQue dataset for GPT-3.5 and GPT-4, respectively (see Table \ref{tab:mhqa_results}).
    \item
    \textcolor{\updatedText}{Using our query-focused hyper-relational KG we use up to $67\%$ fewer tokens to represent query relevant information than the current SoTA method, and up to $60\%$ fewer tokens w.r.t. original supporting documents (see Table \ref{tab:token_count})}.
\end{itemize}

\section{Related Work}
\vspace{-0.1cm}

\textbf{Multi-hop Question Answering (MHQA)}  \space Recent achievements in addressing straightforward questions~\cite{lan2019albert} coupled with  availability of high-quality MHQA datasets~\cite{yang2018hotpotqa, trivedi2022musique} have prompted a shift towards tackling multi-hop questions.

\noindent Existing works in MHQA have adopted various approaches, including: (i) Construction of dynamic graphs ~\cite{qiu-etal-2019-dynamically}, hierarchical ~\cite{fang-etal-2020-hierarchical} and Abstract meaning representation (AMR) based graphs~\cite{deng2022interpretable} as well as leveraging KGs~\cite{li2023leveraging} (ii) Employing end-to-end differentiable learning methods, utilizing Multi-task Transformers for retrieval, reranking and predicting in an iterative manner ~\cite{qi-etal-2021-answering} or using Recurrent Neural Networks (RNN) to sequentially retrieve documents from a graph of entities \cite{DBLP:conf/iclr/AsaiHHSX20} and (iii) dynamically converting multi-hop questions into single-hop questions by generating subsequent questions based on the answers to previous ones~\cite{perez-etal-2020-unsupervised}, or by updating single-hop questions in the embedding space~\cite{sun2021iterative}.  In contrast to these methodologies, our work introduces a novel approach by constructing a hyper-relational KG from the documents, which is then utilized exclusively to answer questions. 



\noindent \textbf{Relational Graphs for MHQA} \space
While retrieving and relying on purely unstructured text is a go to approach for single-hop question answering \cite{watanabe2017question}. It may not be ideal for MHQA, as it is designed to make multi-step, comprehensive reasoning difficult. This issue is often addressed by constructing structured sources of information from raw text and in most of the cases using KGs \cite{li2023leveraging}. By using KGs, the relational information among question concepts and answers can be easily captured \cite{10.1145/3543507.3583376}. However, one of the main drawback of KGs is the lack of context \textit{i.e.}, they focus only on triples, overlooking qualifiers important for inference. To avoid overlooking qualifiers, our method involves building hyper-relational KGs  to serve as an input to LLMs for MHQA, as explained in detail in Sect \ref{subsec:HyperKGConstruction}. To the best of our knowledge, our work is the first to use hyper-relational KGs for MHQA.

\noindent \textbf{Training Free MHQA } \space  
Previously, KGs have been applied to MHQA in two ways: training-based \cite{sun2018open, sun2019pullnet, yavuz2022modeling, ramesh2023single} and training-free \cite{li2023leveraging}, also known as prompting. In these methods, KGs used were either human-curated \cite{speer2017conceptnet, bollacker2008freebase} or training-based \cite{izacard2021leveraging, bosselut2019comet}. In some cases, LLMs were employed for triple extraction from documents to form graphs \cite{li2023leveraging, carta2023iterative}. However, no prior works automatically create a schema for the graph and use the graph for MHQA, a unique approach used by us.





\begin{figure*}
  \centering
  \footnotesize
    \includegraphics[width=0.88\textwidth]{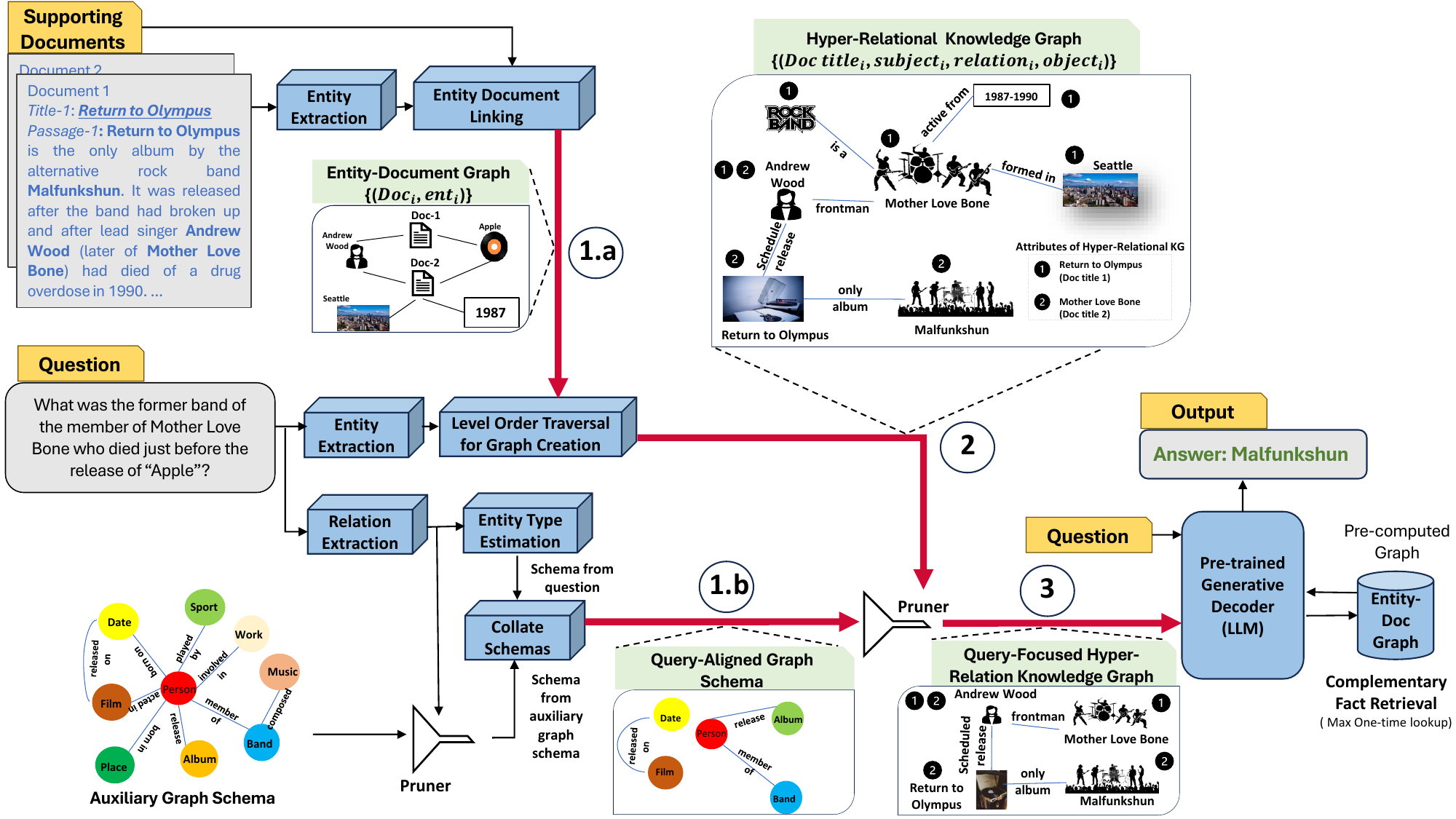} 
\caption{\textbf{Method Overview}: Our method has three key steps - (i) Query-Dependent Structured Knowledge Discovery (Section \ref{subsec:HyperKGConstruction}), (ii) Knowledge Schema Construction for Information Refinement (Section \ref{subsec: knowledge schema construction}), and (iii) Reader LLM Prompt Construction (Section \ref{sec:input_prompt}). Step (i) involves creation of an entity document graph ({\Large{$\textcircled{\small{\textbf{1.a}}}$}} in the Figure), and performing a level-order traversal on it to get a Hyper-relational KG ({\Large{$\textcircled{\small{\textbf{2}}}$}} in the Figure). Next, in step (ii), we create a query-aligned knowledge schema from the question and an auxiliary graph schema ({\Large{$\textcircled{\small{\textbf{1.b}}}$}} in the Figure), and use it to prune the Hyper-Relational KG ({\Large{$\textcircled{\small{\textbf{3}}}$}} in the Figure) - which forms the input for the LLM.}
  \label{fig:approach}
\end{figure*}

\vspace{-0.1cm}
\section{Preliminaries}
\vspace{-0.1cm}
\label{sec: prelim}

\textbf{Hyper-Relational Knowledge Graph} - 
A hyper-relational KG $\mathcal{H}$ is an enriched form of a traditional KG. It allows for the representation of multiple relationships between entities. 

\noindent Let $\mathcal{A}$ denote a collection of \emph{attribute sets} whose elements represent different types of attribute sets. For example, $A_{ts} \in \mathcal{A}$ is a set of \emph{timestamps}, while $A_{dt} \in \mathcal{A}$ is a set of \emph{document titles}, that serve as additional attributes. We then define our hyper relational KG $\mathcal{H}$ as:  
$\mathcal{H} = \{(e_s, r, e_o, a) \mid e_s, e_o \in E, r \in R, a \in A_{dt} \}$,
where $E$ is the set of named entities, $R$ is the set of relations and $A_{dt}$ is the set of document titles that are additional attributes.
This structure not only links entities $e_s$ and $e_o$ via relation $r$ but also integrates the corresponding document title attributes $A_{dt}$, offering a more nuanced description. We refer $(e_s, r, e_o, a)$ as a hyper triple.

\noindent \textbf{Graph Schema} - 
A graph schema $\mathcal{G}_S$, defined as $\mathcal{G}_S = \{(\Tilde{e}_s, r, \Tilde{e}_o) \mid \Tilde{e}_s, \Tilde{e}_o \in E_T, r \in R\}$, outlines the structure of a KG through entity types $E_T$ and relations $R$. Each triple in the schema specifies the permissible entity types $\Tilde{e}_s$ (subject) and $\Tilde{e}_o$ (object) for each relation $r$, serving as a blueprint for organizing knowledge.

\noindent \textbf{Named Entity and Relation Extraction} - 
Named entity extraction identifies textual spans referring to specific entities, while relation extraction classifies semantic relationships between entities in text, enabling structured representation of unstructured data. In this work, we harness LLMs' capabilities for these tasks. Pre-trained on extensive text, these models accurately classify named entities and extract relations with minimal examples \cite{li2023leveraging, wang2023gpt, wadhwa2023revisiting}, showcasing their proficiency in contextual understanding. See Appendices \ref{sec:prompt_enty_extr_supp_docs}, \ref{sec:prompt_enty_rel_extr_query} for detailed prompts and few-shot examples used in our method for named entity and relation extraction.

\vspace{-0.15cm}
\section{Methodology}
\label{sec: method}
The key idea of our method is to identify the subset of documents that contain the answer to the multi-hop query, and subsequently extract context-aware structured information from them (where context comes from the documents). We further perform a refinement step to retain query-relevant information. Below, we begin with a brief description of the problem statement in Section \ref{sec:problem_statement}, followed by an explanation of our method in Section \ref{sec:method_details}. The overview of our approach can be seen in Figure \ref{fig:approach}.

\vspace{-0.1cm}
\subsection{Problem Statement}
\label{sec:problem_statement}
In this work, we address the challenge of MHQA in a zero-shot setting. This implies that we assume no prior domain-specific labeled data, rendering the problem both challenging and practically useful \cite{zero_shot_nlp}. Moreover, we operate in a training-free setting, leveraging the reasoning capabilities of LLMs for this task.

\noindent Formally, the task is to extract the answer $a_q$ from a given natural language question $q$ and a collection of supporting documents $\mathcal{S}_q$. The set of supporting documents is defined as $\mathcal{S}_q \coloneqq \{D^{1}_q, D^{2}_q, \ldots, D^{m}_q\}$, where each document $D^{j}_q \coloneqq (t^{j}_q, p^{j}_q)$ consists of a title $t^{j}_q$ and the associated passage $p^{j}_q$. Given that $q$ is a multi-hop question, deducing its answer requires the aggregation of information from at least two documents \(D^{i}_q, D^{j}_q \in \mathcal{S}_q\) (where \(i \neq j\)).

\vspace{-0.15cm}
\subsection{Proposed Method (\name)} 
\label{sec:method_details}
We begin by traversing the supporting documents to identify the subset of documents relevant to the query, and extract structured, context-aware information (Section \ref{subsec:HyperKGConstruction}). This information is then refined using a query-based knowledge schema (Section \ref{subsec:prunning}). 
Next, we format the distilled graph for the reader LLM prompt, including a complementary fact retrieval step within the prompt to ensure higher coverage of query-relevant information.
(Section \ref{sec:input_prompt}).

\vspace{-0.2cm}
\subsubsection{Query-Dependent Structured Knowledge Discovery}
\label{subsec:HyperKGConstruction}
Unstructured text, contains a complex web of facts and relationships forming a latent semantic graph. This graph represents interconnected factual information not explicitly structured but implied by relationships and named entities in the text. Our approach navigates this structure using the entity-document graph—a bipartite graph linking documents and named entities. It helps uncover parts of the latent graph relevant to the given query.


\noindent \textbf{Entity-Document Graph Construction} The entity-document graph has two node types—documents and entities—with a single edge connecting them. We begin by extracting named entities from supporting documents. Then, we establish edges between document and entity nodes, forming a bipartite graph that captures the connections between entities and the documents they appear in (See Figure \ref{fig:approach} - {\Large{$\textcircled{\small{\textbf{1.a}}}$}} for an example).

\vspace{0.15cm}
\noindent \textbf{Level Order Traversal and Structured Information Extraction}
We begin by extracting named entities from the query and use them for a level-order/breadth-first traversal of the entity-document graph. Similar to \citet{li2023leveraging, wadhwa2023revisiting}, we use LLMs to extract KG triples from document nodes, framing the task as a sequence-to-sequence problem. By providing the LLM with a detailed prompt (a clear description of the task and output format) and few-shot examples, we guide the model to generate triples directly from raw text (see Appendix \ref{sec:prompt_triples_extr_docs} for prompts).

\noindent After extracting triples, we enhance them into hyper-relational KG quadruples (or hyper triples) by appending the title of the source document to each triple. This title serves as an additional attribute, offering the LLM context to understand when the triple is valid. 
During the traversal, we filter hyper triples with the current entity node as either the subject or object, adding the counterpart entity to the traversal queue. This iterative process continues for a predefined number of hops $k$, where the value of $k$ defines the query complexity, \textit{i.e.}, the number of facts required to answer the question. Example of a hyper-relational KG in Figure \ref{fig:approach}-{\Large{$\textcircled{\small{\textbf{2}}}$}}.

\noindent When we perform the above specified $k$-hop traversal starting from the named entities in the query, we are effectively probing the latent semantic graph in the raw text by incrementally 
selecting subsets of documents related to the query.

\subsubsection{Knowledge Schema Construction for Information Refinement}
\label{subsec: knowledge schema construction}
The hyper-relational graph construction is rooted in the inference query. However, it contains some hyper triples which although are related to the named entities in the query, capture relationships that are not useful for answering it. For instance, a query might focus on a person's occupation, but the graph also captures their hobbies. To eliminate such extraneous information, we construct a query-aligned knowledge schema (Figure \ref{fig:approach} - {\Large{$\textcircled{\small{\textbf{1.b}}}$}} shows an example) and then carry out a refinement step (refined output example in Figure \ref{fig:approach} - {\Large{$\textcircled{\small{\textbf{3}}}$}}).

\noindent \textbf{Query-Aligned Graph Schema Creation}
\label{subsec:schema_creation}
The graph schema acts as a structural template, guiding the organization of information crucial for answering the given question. It is essential for capturing the types of entities and relations likely to form the backbone of the answer. For example, in response to the question \textit{"Who is the 2nd daughter of the 1st president of X?"} the schema should capture relations such as \{<Person>, daughter of, <Person>\} and \{<Person>, president of, <Country>\}, representing the direct information sought by question.

\noindent Our graph schema is populated using two sources:

\vspace{0.05cm}
\noindent (i) We first derive schema elements from the inference query by identifying relations in it, and then using LLMs to estimate the subject and object entity types for each of these relations (prompt in Appendix \ref{sec:prompt_enty_type_est}). This forms a schema with a direct alignment with the question's intent.

\vspace{0.05cm}
\noindent (ii) We enrich the schema with additional domain-specific relations to aid multi-hop reasoning, using an auxiliary graph schema derived from in-domain questions (Appendix \ref{sec:auxiliary_graph_schema_creation}). This includes a wider array of entity-relation pairs, such as \{<Person>, child of, <Person>\} and \{<Person>, head of, <Country>\}, to account for potential inferential steps when direct query matches are absent in supporting documents. We select schema triples with relations similar to the question's (based on cosine similarity) to retrieve a relevant subset. While constructing this auxiliary schema incurs a one-time compute cost, its amortized cost is minimal over multiple queries, especially given it helps to reduce the input token length for the reader LLM by upto 60\% w.r.t. original source documents.

\noindent In cases where the inference query diverges from the auxiliary graph schema, we rely on the schema derived from the query. This ensures system flexibility, adapting to specific query requirements, while still benefiting from the broader knowledge in the auxiliary schema when applicable.

\vspace{0.1cm}
\noindent \textbf{Pruning Hyper-relational Knowledge Graph}
\label{subsec:prunning}
The final step in our methodology involves refining the constructed hyper-relational KG `$\mathcal{H}$' by pruning it using the query-aligned graph schema $\mathcal{G}_S$. This pruning process is essential to distill the graph to its most relevant components, thereby enhancing the efficiency and effectiveness of the reader LLM in generating answers.

\noindent The pruning process begins by computing embeddings for the relations in both the hyper-relational KG and the graph schema. Let $\mathbf{v}_r$ denote the embedding of a relation $r$ in the hyper-relational KG, and let $\mathbf{u}_{r'}$ denote the embedding of a relation $r'$ in the graph schema. We compute cosine similarity between each pair of relation embeddings:
\begin{equation}
\text{sim}(\mathbf{v}_r, \mathbf{u}_{r'}) = \frac{\mathbf{v}_r \cdot \mathbf{u}_{r'}}{\|\mathbf{v}_r\| \|\mathbf{u}_{r'}\|},
\end{equation}

\noindent where $\cdot$ denotes the dot product and $\|\cdot\|$ denotes the Euclidean norm. For each relation $r_h$ in a hyper triple $(e_s, r_h, e_o, a) \in \mathcal{H}$, we compute its highest similarity score w.r.t. any relation $r_s$ in the schema:
\begin{equation}
\text{score}(r_h) = \max_{r_s \in \mathcal{G}_S} \text{sim}(\mathbf{v}_{r_h}, \mathbf{u}_{r_s}).
\end{equation}
We compute the scores for all those relations whose entity types in the hyper triple match with that in the schema. We then select $p$ hyper triples with the highest scores. This gives us the pruned graph $\mathcal{H}' = \text{sort}_{\text{score}}(\mathcal{H}, p)$, where $\text{sort}_{\text{score}}$ denotes the sorting operation based on the computed similarity scores, and $p$ - no. of hyper triples to retain.

\subsubsection{Reader LLM Prompt Construction}
\label{sec:input_prompt}
Next, we create the prompt for the reader LLM. Each hyper triple in the pruned hyper-relational KG, $\mathcal{H}'$, is verbalized into an English sentence as the LLMs are adept at understanding the same \cite{jiang2023structgpt}. The verbalization process transforms the structured triple into a natural language text. To be specific, we concatenate each item in the hyper triple into a long sentence marked by specific separation and boundary symbols.
The resulting sentences are then arranged in the descending order of their similarity scores w.r.t. the schema to form the input prompt. We arrange the facts based on the relevance to the query (measured via similarity scores) as the information retrieval by LLMs (w.r.t. their input prompt) is done best when the gold information is placed closest to query \cite{cuconasu2024power}. \\
As structured information extraction is an unsolved problem, some pertinent details may be missed in the input graph. To mitigate this, we include a verification step in the prompt, described below. 

\vspace{0.1cm}
\noindent \textbf{Complementary Fact Retrieval}
\label{sec:comp_fact_retrieval}
If the LLM identifies that facts about a particular set of named entities is missing from the input graph, then we instruct it to list those named entities.
We then fetch the corresponding documents from the entity-document graph and integrate them with the initial set of relevant facts. This process not only enriches the input for the LLM but also ensures that any missing query-relevant information is retrieved, enhancing the accuracy of the system's responses. We assume that a single-step verification is sufficient. \\
Refer Appendix \ref{sec:prompt_reader_LLM} for the reader LLM prompt.

%

\section{Experimental Setup}
\label{sec:exp_setup}
\subsection{Evaluation Details}
\noindent \textbf{Datasets}:
\label{sec:exp_dataset}
We use two benchmark multi-hop question-answering datasets, namely HotpotQA \cite{yang2018hotpotqa} and MuSiQue \cite{trivedi2022musique} for evaluating our method. 
Table \ref{tab:QA_stats} displays the total number of samples for both datasets in the training and development sets.

\noindent For our evaluation process, we utilize questions, context, and gold answers from the development sets of both HotpotQA and MuSiQue. To tune hyperparameters, we randomly select 50 questions from the development set and use them (following \cite{trivedi2022interleaving,li2023leveraging}). We create our test set by randomly sampling an additional 1000 questions from the development set of HotpotQA and 1200 questions from development set of MuSiQue. Its worth noting that our test set size is twice as big as StructQA \cite{li2023leveraging}. 
In both HotpotQA and MuSiQue datasets, for each question there are ten and twenty supporting documents available, respectively. Each document in both datasets contains a title and a passage of text.


\renewcommand{\arraystretch}{1.1} 
\begin{table*}[t]
\footnotesize
\captionsetup{font=footnotesize}
\centering
\begin{tabular}{lcccccccc}
\cline{1-9}     
\multicolumn{1}{l}{\textbf{Datasets}} 
& \multicolumn{4}{c}{HotpotQA} 
& \multicolumn{4}{c}{MuSiQue} 
\\
\cmidrule(lr){2-5} \cmidrule(lr){6-9} 
\multicolumn{1}{l}{\textbf{ Methods}}
& \multicolumn{1}{p{1.1cm}}{\centering EM ($\uparrow$)} 
& \multicolumn{1}{p{1cm}}{\centering F1 ($\uparrow$)} 
& \multicolumn{1}{p{1cm}}{\centering P ($\uparrow$)}
& \multicolumn{1}{p{1cm}}{\centering R ($\uparrow$)}
& \multicolumn{1}{p{1.1cm}}{\centering EM ($\uparrow$)} 
& \multicolumn{1}{p{1cm}}{\centering F1 ($\uparrow$)} 
& \multicolumn{1}{p{1cm}}{\centering P ($\uparrow$)} 
& \multicolumn{1}{p{1cm}}{\centering R ($\uparrow$)}
\\ 
\cline{1-9}
\cline{1-9}
\rowcolor{gray!15} 
\multicolumn{9}{c}{\centering \fontfamily{lmss}\selectfont{ \textit{Reader: gpt-4-1106-preview}}} \\
\cline{1-9} 
Base (w/o supp docs) & 0.26 & 0.45 & 0.45 & 0.50 & 0.09 & 0.21 & 0.22 & 0.21
\\
Base (with supp docs) & 0.54 & 0.74 & 0.75 & 0.77 & 0.39 & 0.55 & 0.55 & 0.56
\\
StructQA \cite{li2023leveraging} & 0.55 & 0.77 & 0.75 & \textbf{0.80} & 0.42 & 0.56 & 0.57 & 0.56
\\
\rowcolor{blue!10}
Our Method & \textbf{0.66} & \textbf{0.78} & \textbf{0.80} & 0.79 & \textbf{0.48} & \textbf{0.58} & \textbf{0.59} & \textbf{0.59}
\\
\cline{1-9}
\cline{1-9}
\rowcolor{gray!15} 
\multicolumn{9}{c}{\centering \fontfamily{lmss}\selectfont{ \textit{Reader: gpt-3.5-turbo-1106}}} \\
\cline{1-9}

Base (w/o supp docs) & 0.23 & 0.37 & 0.38 & 0.40 & 0.06 & 0.15 & 0.17 & 0.15
\\
Base (with supp docs) & 0.47 & 0.65 & 0.66 & 0.68 & 0.24 & 0.36 & 0.36 & \textbf{0.37}
\\
StructQA \cite{li2023leveraging} & 0.48 & 0.64 & 0.62 & 0.67 & 0.23 & 0.37 & 0.37 & \textbf{0.37}
\\
\rowcolor{blue!10}
Our Method & \textbf{0.57} & \textbf{0.69} & \textbf{0.69} & \textbf{0.70} & \textbf{0.29} & \textbf{0.38} & \textbf{0.39} & \textbf{0.37} \\
\cline{1-9}
\end{tabular}
\caption{\textbf{Multi-hop Reasoning Evaluation (Automatic Metrics)}: We report the Exact-Match (EM), F1, Precision (P) and Recall (R) scores of all methods in comparison on two MHQA datasets. We experiment with two SoTA reader LLMs for the QA task - GPT-4 and GPT-3.5. We report results on Gemini-pro in the Appendix \ref{sec:appx_further_res_anal}. The results indicate consistent and significant improvements across datasets, metrics and LLMs. \textit{Base}: Only reader LLM; \textit{supp docs}: supporting documents w.r.t. query}
\label{tab:mhqa_results}
\end{table*}

\vspace{0.1cm}
\noindent \textbf{Baselines}:
We operate in a training-free setting, utilizing LLMs as the \textit{reader} model (takes the question and corresponding supporting documents as input and gives an answer). We experiment with two popular LLMs - GPT-3.5, GPT-4 \cite{achiam2023gpt} in the main paper, and report results with Gemini \cite{lee2023gemini} in Appendix \ref{sec:appx_further_res_anal}. 
We compare our method against three baselines\\ (i) \textit{StructQA} \cite{li2023leveraging}:  CoT \cite{wei2022chain} based SoTA method for MHQA with LLMs. We use the same prompts and methodology as outlined in their study. \\ (ii) \textit{Base (with supp docs)}: To test the base multi-hop reasoning of the reader LLM, in this baseline, we feed the reader LLM with the question and supporting documents directly. \\(iii) \textit{Base (w/o supp docs)}: To test the parametric knowledge of the reader LLM, in this baseline, we feed it with just the question and elicit a response. \\
All prompts used in our experiments are documented in the Appendix \ref{sec:appx_prompt} for reproducibility.





\vspace{0.15cm}
\noindent \textbf{Evaluation Metrics}:
We use Exact-Match (EM), Precision, Recall, and F1-Score as automatic metrics \cite{li2023leveraging, mavi2022survey} to measure correctness of the predicted answers. \\
\noindent For semantic evaluation of the predicted answers, we also compute Human Evaluation Score (H-Eval) and BERTScore \cite{zhang2019bertscore} on a random sample of 100 questions from the development set of the HotpotQA dataset. The Human Evaluation Scores are obtained by averaging scores from three annotators. We use a smaller subset of question for this analysis due to the resource-intensive nature of human evaluation.

\vspace{-0.2cm}
\subsection{Implementation Details}
\noindent \textbf{Problem Setting} We focus on the \textit{distractor} setting \cite{yang2018hotpotqa, trivedi2022musique} for question answering. In this setting, the supporting documents for each question consist of a set of distractor documents, \textit{i.e.}, documents not useful for predicting the answer. This setting poses a challenge, demanding robustness to noise in input. 

\vspace{0.1cm}
\noindent \textbf{Knowledge Triple Extraction} Both StructQA and our method use LLMs for knowledge triple extraction. Thus, for a fair comparison, we always use the same triple extractor LLM for both. Results in Table \ref{tab:mhqa_results}, use \texttt{gpt-3.5-turbo-instruct} for triple extraction. We further study the impact of different triple extractor LLM in Table \ref{tab:triple_extractor_sensitivity}.

\vspace{0.1cm}
\noindent \textbf{\name hyperparameters} We use OpenAI embedding model (\texttt{text-embedding-ada-002}) to compute text embeddings. We set the value of $k$ (number of levels in the level order traversal) to $4$ across both the datasets in Table \ref{tab:mhqa_results},\ref{tab:triple_extractor_sensitivity}. Beyond $4$ levels of traversal, performance remains the same as the complexity of the dataset vary from 2-4 hop (See Table \ref{tab:QA_stats_hop-wise}). Similarly, we set the value of $p$ (number of hyper triples to be retained after pruning) to $50$ across all result tables
(sensitivity analysis in the Appendix \ref{sec:appx_further_res_anal}). Both of these values were chosen by experimenting on $50$ samples from the development sets of the respective datasets. 


\noindent \textbf{Auxiliary Graph Schema Creation} We randomly sample 10,000 questions (without answers and supporting documents) from the training data of HotpotQA and MuSiQue, for creating the graph schema. We provide further details about the auxiliary schema creation in the Appendix \ref{sec:auxiliary_graph_schema_creation}.

\vspace{-0.1cm}
\section{Results and Analysis}
\label{sec:results}
\vspace{-0.1cm}


Here, we study the performance of our method by investigating several key dimensions of multi-hop question answering (especially in the era of LLMs):
\begin{enumerate}[label=(\roman*),nosep]
    \item Multi-hop reasoning capability
    \begin{enumerate}[nosep]
        \item w.r.t automatic metrics
        \item w.r.t human \& semantic metrics
    \end{enumerate}
    \item \textcolor{\updatedText}{Performance w.r.t. different question types}
    \begin{enumerate}[nosep]
        \item \textcolor{\updatedText}{Reasoning type-wise performance}
        \item Hop-wise performance
    \end{enumerate}
    \item \textcolor{\updatedText}{Query Information efficiency (input token count for reader LLM)}
    \item Measure of confident predictions
\end{enumerate}
We also study the impact of the LLM used for knowledge triple extraction, and conduct further studies in the App \ref{sec:appx_further_res_anal} and a case study in App \ref{sec:appx_case_study}.




\vspace{0.1cm}
\noindent \textbf{Multi-hop Reasoning}:
\noindent \textbf{(a) Evaluation using Automatic Metrics} - In Table \ref{tab:mhqa_results} we report the  performance of our method w.r.t. Exact Match (EM) and F1 scores for both datasets. We find that, across reader LLMs, our method consistently outperforms all baseline methods. This underscores the importance of our data organization and pruning process, retaining only the relevant information in input prompts for the reader LLMs. Notably, our method even outperforms the SoTA StructQA, which employs the CoT mechanism \cite{wei2022chain}. 



\noindent \textbf{(a) Human and Semantic Evaluation}
As generative models can generate long worded answers, its important to semantically evaluate the predicted answers. Thus, we use BertScore and human evaluators to judge the correctness of the predicted answers. We use GPT-4 reader based responses for this study and report the results in Table \ref{tab:human_eval}.

\begin{table}
\renewcommand{\arraystretch}{1.15} 
\centering
\footnotesize
\captionsetup{font=footnotesize}
\vspace{-0.3cm}
\hspace{-0.2cm}
\begin{tabular}{lcc}
\toprule
\textbf{Dataset} & \multicolumn{2}{c}{HotpotQA} \\
\cmidrule(lr){2-3}
\textbf{Methods} & \multicolumn{1}{p{1.4cm}}{\centering H-Eval ($\uparrow$)} & \multicolumn{1}{p{1.5cm}}{\centering B-Score ($\uparrow$)} \\
\midrule
Base & 86.5 & 87.00 \\
StructQA & 87.5 & 85.20 \\
\rowcolor{blue!10}
Our Method & \textbf{89.0} & \textbf{89.05} \\
\bottomrule
\end{tabular}
\vspace{-0.1cm}
\caption{\textbf{Multi-hop Reasoning Evaluation (Semantic Metrics):} Human evaluation (H-Eval) on 100 samples \& BERTScore (B-Score) for 1000 samples from HotpotQA.}
\label{tab:human_eval}
\vspace{-0.25cm}
\end{table}

We observe that our method achieves a 1-2 point improvement on both the metrics w.r.t. the baselines, confirming the efficacy of our method.

\vspace{0.1cm}
\noindent \textbf{Query Information Efficiency}
In multi-hop question answering, where reader LLMs like GPT-4 are employed, the presence of irrelevant information poses a significant challenge because (i) it increases the computational load (and thus the cost) (ii) and complicates the LLM's task of connecting disparate pieces of information across documents. Thus, efficiently managing input data by filtering out unnecessary content becomes crucial to both performance and cost-effectiveness. To quantify this factor, we measure the reader input token count and use an efficiency metric for input tokens. The query information effiency (or reader input token efficiency) metric is a normalized score between zero and one, computed through min-max normalization of input token count (min \& max are based on input token counts of all methods in comparison). Our results, in Figure \ref{fig:radar_plot} and Table \ref{tab:token_count}, show a significant reduction in average input token count across both datasets compared to baselines. 

\renewcommand{\arraystretch}{1.1} 
\begin{table}[!ht]
    \centering
    \vspace{-0.2cm}
    \footnotesize
    \captionsetup{font=footnotesize}
    \begin{tabular}{lcc}
    \hline
        \textbf{Datasets} & HotpotQA & MuSiQue \\ 
        \textbf{Methods} & Token Count($\downarrow$) & Token Count($\downarrow$) \\ \hline
        \textcolor{gray!55}{Supp docs}  & \textcolor{gray!55}{1333.81}  & \textcolor{gray!55}{2361.36}  \\ 
        StructQA & 3078.85  & 5908.87  \\
        \rowcolor{blue!10}
        Our Method & \textbf{1230.90}  & \textbf{1398.15}  \\ \hline
    \end{tabular}
    \caption{\textbf{Reader Input Token Count}: Comparing the average input token size across datasets to measure the efficiency of representing query relevant information.}
    \label{tab:token_count}
    \vspace{-0.2cm}
\end{table}

\noindent \textcolor{\updatedText}{
\textbf{Performance w.r.t. different question types}\\ To analyze \name's performance across question types and complexities, we perform an analysis based on reasoning types (with HotpotQA dataset) and hops (with MuSiQue dataset). We choose respective datasets for each analysis based on the query type information available in the datasets.}

\noindent \textcolor{\updatedText}{
\textbf{(a) Reasoning type-wise performance} \\
For any multi-hop question, the data generating process dictates the type of reasoning required to answer the question. HotpotQA dataset contains questions which fall in two categories w.r.t. reasoning - (i) bridge question (ii) comparison questions. We advise the reader to refer to the dataset paper \cite{yang2018hotpotqa} for more details. We report results across question categories in Table \ref{tab:reasoning_type_bridge_comparison}.}

\begin{table}[!ht] 
\renewcommand{\arraystretch}{1.15} 
\centering
\footnotesize
\captionsetup{font=footnotesize}
\vspace{-0.2cm}
\hspace{-0.2cm}
\begin{tabular}{l@{\hskip 0.2cm}c@{\hskip 0.2cm}c@{\hskip 0.2cm}c@{\hskip 0.2cm}>{\columncolor{blue!10}}c}
\toprule
\textbf{Type} & \textbf{\# Samples} & \textbf{Base} & \textbf{StructQA} & \textbf{\name} \\
\midrule
Bridge & 787 & 0.55 & 0.56 & \textbf{0.64} \\
Comparison & 213 & 0.55 & 0.51 & \textbf{0.71} \\
\bottomrule
\end{tabular}
\vspace{-0.1cm}
\caption{\textbf{Performance Across Reasoning Types:} Evaluation of different methods. \textit{Base} - Reader LLM with supp docs.}
\label{tab:reasoning_type_bridge_comparison}
\vspace{-0.25cm}
\end{table}
\vspace{0.1cm}
\noindent \textbf{(b) Hop-wise Performance Comparison} \\
\vspace{-0.3cm}

\begin{figure}[!ht]
  \centering
  \footnotesize
  \captionsetup{font=footnotesize}
  \includegraphics[width=0.3\textwidth]{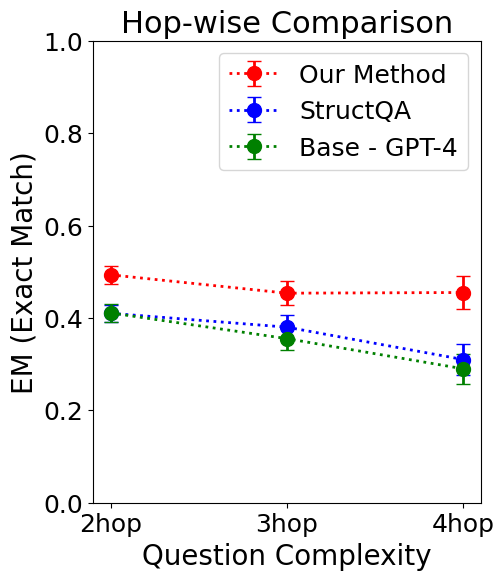}
  \caption{\textbf{Hop-wise Performance:} Comparison on MuSiQue dataset with varying question complexity. Bars denote standard error - $\sqrt{\frac{em (1 - em)}{n}}$.} 
  \label{fig:hop_wise_perf_graph}
  \vspace{-0.4cm}
\end{figure}
\noindent We evaluated the hop-wise performance of our method on MuSiQue dataset (Figure \ref{fig:hop_wise_perf_graph}). This evaluation aimed to determine if our performance improvements (in Table \ref{tab:mhqa_results}) were primarily attributed to the system's ability to answer 2-hop questions, which are less complex than 3 and 4-hop questions. We find that our method maintains its performance across an increasing number of hops and outperforms the baselines too.

\noindent \textbf{Measure of Confident Predictions}
LLMs demonstrate reasoning capabilities but often struggle to recognize their limitations in understanding queries or retrieving correct answers. We found that structured inputs and task-dependent reasoning steps, including the option for models to indicate uncertainty, alleviates this issue. Inspired by this, we introduce the \textit{Self-Aware Exact Match} (Self-Aware EM) score to assess LLMs on complex QA tasks. This metric evaluates the accuracy and confidence of model responses, focusing on instances where the model provides answers with high confidence (or self-awareness). It aims to highlight the model's precision and reliability, offering a nuanced view of its performance.
It is defined as:
$\text{Self Aware EM} = \frac{\sum_{q \in Q_A} \text{EM}(q)}{|Q_A|}$
where $Q_A$ is the set of questions the system answers. We report the Self-Aware EM scores in Figure \ref{fig:radar_plot}. \\
In our evaluation using the HotpotQA and MuSiQue datasets, \name\ opted for "None" or "No answer" in 3\% and 10\% of the samples, respectively (with GPT-4 as reader LLM). In contrast, StructQA always provided an answer, often inaccurately (see Table \ref{tab:mhqa_results}). Upon reviewing instances where \name\ did not provide an answer, we found it was due to incomplete information in the input graph or a misunderstanding of the question. This self-awareness—recognizing when it cannot reliably answer—is a key feature of dependable systems. Our findings, detailed in Figure \ref{fig:radar_plot}, show that \name\ achieves significantly higher self-aware EM scores compared to other baselines.

\vspace{-0.1cm}
\section{Sensitivity \& Ablation Analysis}
\vspace{-0.3cm}
\begin{table}[!ht]
\renewcommand{\arraystretch}{1.15} 
\centering
\footnotesize
\captionsetup{font=footnotesize}
\vspace{-0.2cm}
\hspace{-0.2cm}
\begin{tabular}{@{\hskip 0.1cm}c@{\hskip 0.1cm}c@{\hskip 0.1cm}c@{\hskip 0.1cm}c@{\hskip 0.1cm}c@{\hskip 0.1cm}c@{\hskip 0.1cm}c@{\hskip 0.1cm}c}
\toprule
\multicolumn{4}{c}{\textbf{Configuration}} & \multicolumn{3}{c}{\textbf{Performance}}\\
\cmidrule(lr){1-4} \cmidrule(lr){5-7}
\makecell{\textbf{Hyper} \\ \textbf{KG}} & \textbf{Prune} & \makecell{\textbf{Aux.} \\ \textbf{Schema}} & \makecell{\textbf{Comp. Fact} \\ \textbf{Retrieval}} & \textbf{EM} & \textbf{F1} & \makecell{\textbf{Reader Inp} \\ \textbf{Tokens}} \\
\midrule
X & - & - & - & 0.56 & 0.69 & 1867 \\
X & X & - & - & 0.56 & 0.70 & 1227 \\
X & X & X & - & 0.58 & 0.72 & 1220 \\
X & X & X & X & 0.61 & 0.77 & 1230 \\
\bottomrule
\end{tabular}
\vspace{-0.1cm}
\caption{\textcolor{\updatedText}{\textbf{Impact of different components:} Performance metrics (EM, F1) and Reader Input Token Count for different configurations of \name. \textbf{X} refers to "\textit{included}", and \textbf{-} refers to "\textit{excluded}" in the table. For example, in the first row, only Hyper-Relational KG is included in the HOLMES algorithm, every other component is removed. These results are reported on a set of 100 samples from HotpotQA dev set.}}
\label{tab:ablation_study}
\vspace{-0.25cm}
\end{table}

\noindent \textcolor{\updatedText}{
\textbf{Ablation Studies}
In Table \ref{tab:ablation_study} we share the results of our ablation study that systematically evaluates the contribution of each component within the \name framework.}

\renewcommand{\arraystretch}{1.15} 
\begin{table}[h]
\footnotesize
\captionsetup{font=footnotesize}
\begin{tabularx} {\columnwidth}{ccccc}
\cline{1-5} 
\multicolumn{1}{c}{\textbf{ \centering Datasets }} 
& \multicolumn{4}{c}{HotpotQA \scriptsize{(100 samples)}} 
\\
\cmidrule(lr){2-5} 
\multicolumn{1}{c}{\textbf{ Methods }}
& \multicolumn{1}{p{1.1cm}}{\centering EM ($\uparrow$)} 
& \multicolumn{1}{p{1cm}}{\centering F1 ($\uparrow$)} 
& \multicolumn{1}{p{1cm}}{\centering P ($\uparrow$)}
& \multicolumn{1}{p{1cm}}{\centering R ($\uparrow$)}
\\ 
\cline{1-5}
\cline{1-5}
\rowcolor{gray!15} 
\multicolumn{5}{c}{\centering \fontfamily{lmss}\selectfont{ \textit{Triple Extractor: gpt-4-1106-preview}}} \\

\cline{1-5} 
StructQA & 0.56 & 0.76 & 0.79 & 0.77
\\
\rowcolor{blue!10}
Our Method & \textbf{0.68} & \textbf{0.79} & \textbf{0.82} & \textbf{0.80}
\\
    
\cline{1-5}
\rowcolor{gray!15} 
\multicolumn{5}{c}{\centering \fontfamily{lmss}\selectfont{ \textit{Triple Extractor: gpt-3.5-turbo-1106}}} \\
\cline{1-5}
StructQA & 0.47 & 0.71 & 0.72 & 0.76
\\
\rowcolor{blue!10}
Our Method & \textbf{0.61} & \textbf{0.77} & \textbf{0.78} & \textbf{0.78}
\\

\cline{1-5}
\end{tabularx}
\caption{\textbf{Impact of Triple Extractor} on MHQA performance}
\label{tab:triple_extractor_sensitivity}
\vspace{-0.3cm}
\end{table}

\noindent \textbf{Sensitivity Analysis}
We evaluated the effect of using a different LLM (\texttt{gpt-4-1106-preview}) for triple extraction, comparing our performance with StructQA in Table \ref{tab:triple_extractor_sensitivity} (reader: GPT-4). Our approach consistently outperforms StructQA, although improved triple extraction models benefit both.

\section{Conclusion}

We introduce \name, an approach leveraging a hyper-relational KG to enhance multi-hop question answering by minimizing noise and refining relevant facts. Constructing an entity-doc graph from the question's supporting documents, we employ a level-order traversal, prune with an auxiliary graph schema, and utilize distilled graph as input for LLM-based answering. 
\textcolor{\updatedText}{Our method achieves SoTA performance with a 20\% improvement on HotpotQA and 26\% on MuSiQue dataset w.r.t EM
while using upto 67\% fewer tokens to represent query relevant information w.r.t. SoTA methods.}


\vspace{-0.5cm}
\textcolor{\updatedText}{
\section{Acknowledgement}
We would like to thank Shivangana Rawat and other members of the AI lab at Fujitsu Research India for providing valuable feedback on the manuscript. We also thank the anonymous ARR reviewers, the meta-reviewer, and the ACL program chairs for their comments and feedback, which helped improve our draft.
}

\section*{Limitations}

The following details outline areas for future improvements:

\noindent \textbf{Possible Incompleteness in Constructed Graphs}  Extracting entities and triples to construct graphs via LLMs demonstrates high accuracy, yet the task of capturing all relevant triples faces inherent challenges. Structured information extraction remains an unsolved problem across the field, leading occasionally to incomplete graphs in our method as well

\noindent \textbf{Increased Computational Effort}: Our method's use of LLMs to generate an auxiliary schema introduces additional computational steps. While this increases the computational effort compared to prior methods, it's offset by our efficiency w.r.t fewer input tokens during inference. 

\noindent While acknowledging the limitations of our approach, we remain optimistic, recognizing our work as a step forward in enhancing the capabilities of LLMs for multi-hop question answering. 

\section*{Ethical Concerns}
There are no ethical concerns associated with this work.

\bibliography{acl_camera_ready}

\appendix

\newpage
\section{Appendix}
\label{sec:appendix}
In this section, we provide additional results and details that we could not include in the main paper due to space constraints.
In particular, this appendix contains the following:
\begin{itemize}[nosep]
    \item \hyperref[sec:appx_further_res_anal]{Additional Results and Analysis} 
    \item \hyperref[sec:appx_dataset_stats]{Dataset Statistics}
    \item \hyperref[sec:appx_guidelines_human_annot]{Guidelines for Human Annotators}
    \item \hyperref[sec:appx_prompt]{ Prompts used in \name}
    \item \hyperref[sec:auxiliary_graph_schema_creation]{Auxilliary schema construction}
    \item \hyperref[sec:appx_case_study]{Case Study of \name on HotpotQA dataset}
\end{itemize}

\subsection{Additional Results and Analysis}
\label{sec:appx_further_res_anal}
Extending our analysis in Section \ref{sec:results}, here we report the following - (i) MHQA performance with Gemini-Pro reader LLM (ii) Sensitivity Analysis w.r.t. pruning process (iii) Sensitivity Analysis w.r.t. the number of depth of traversal while creating the Hyper-Relational KG.

\vspace{0.2cm}
\noindent \textbf{Gemini-Pro MHQA Results}
\renewcommand{\arraystretch}{1.15} 
\begin{table}[!ht]
\footnotesize
\begin{tabularx} {\columnwidth}{ccccc}
\cline{1-4} 
\multicolumn{1}{c}{\textbf{ \centering Datasets }} 
& \multicolumn{3}{c}{\centering HotpotQA} 
\\
\cmidrule(lr){2-4} 
\multicolumn{1}{c}{\textbf{ Methods }}
& \multicolumn{1}{p{1.1cm}}{\centering EM ($\uparrow$)} 
& \multicolumn{1}{p{1cm}}{\centering F1 ($\uparrow$)} 
& \multicolumn{1}{p{1.5cm}}{\centering SA-EM ($\uparrow$)}
\\ 
\cline{1-4}
\rowcolor{gray!15} 
\multicolumn{4}{c}{\centering \fontfamily{lmss}\selectfont{ \textit{Reader: Gemini-Pro}}} \\

\cline{1-4} 
Base (w/ supp docs) & 0.48 & 0.66 & 0.48
\\
StructQA & 0.49 & 0.66 & 0.52
\\
\rowcolor{blue!10}
Our Method & \textbf{0.58} & \textbf{0.67} & \textbf{0.66} 
\\
\cline{1-4}

\end{tabularx}
\caption{\textbf{Multi-Hop QA performance} \textit{SA-EM}: Self-Aware EM; \textit{Base}: Only reader LLM; \textit{supp docs}: supporting documents w.r.t. query}
\label{tab:gemini_hotpot_results}
\end{table}

\noindent Similar to results with GPT-3.5 and GPT-4 reader LLMs (in Table \ref{tab:mhqa_results}), we observe consistent improvements across metrics. Thus demonstrating the efficacy of our method, across LLMs.

\vspace{0.2cm}
\noindent \textbf{Impact of Pruning} 
\vspace{0.1cm}

\begin{figure}[!ht]
  \centering
\includegraphics[width=0.6\columnwidth]{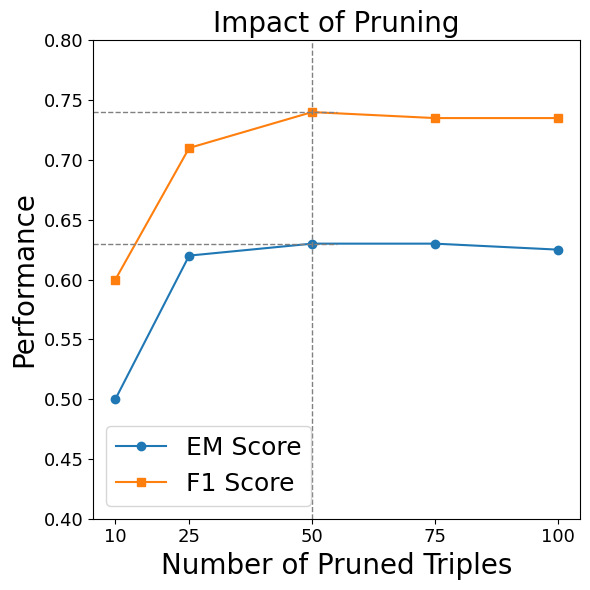}
  \caption{\textbf{Impact of pruning} on MHQA performance in HotpotQA dataset}
  \label{fig:sensitivity_anal_prunning}
\end{figure}

\noindent In Tables \ref{tab:mhqa_results} and \ref{tab:triple_extractor_sensitivity}, we conducted our experiments by retaining $50$ hyper triples after the pruning stage. Here we vary the number of hyper triples being retained after pruning and study the impact on the MHQA performance (EM and F1 scores).

\noindent Figure \ref{fig:sensitivity_anal_prunning} shows that after $50$ triples, the MHQA performance does not change indicating that the first $50$ triples captures all the query relevant information. This number is dependent on the dataset, so we suggest conducting experiments to deduce this threshold. 

\vspace{0.2cm}
\noindent \textbf{Impact of Depth in the Level Order Traversal}

\begin{figure}[!ht]
  \centering
\includegraphics[width=0.6\columnwidth]{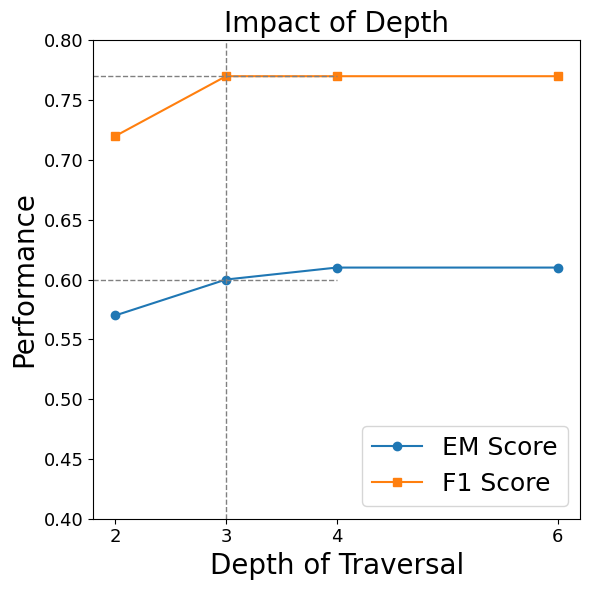}
  \caption{\textbf{Impact of Depth} on MHQA performance in HotpotQA dataset}
  \label{fig:sensitivity_anal_depth}
\end{figure}

\noindent In the Tables \ref{tab:mhqa_results} and \ref{tab:triple_extractor_sensitivity}, we conducted our experiments by traversing the entity document graph upto a depth of four levels. As stated in Section \ref{subsec:HyperKGConstruction}, the depth of traversal depends on the maximum complexity of query we expect to see during inference. To verify this, we perform a sensitivity analysis on this variable. We experiment on $100$ randomly sampled datapoints from HotpotQA development set(question complexity varies between 2-3 hops).

\noindent Figure \ref{fig:sensitivity_anal_depth} reflects that as the max query complexity of the dataset is three, beyond three levels of traversal the performance does not improve. Thus, verifying our claim. We choose a depth of four in the main paper as we wanted to cater to all the datasets.

\noindent \textcolor{\updatedText}{\textbf{Total Input and Output Token Lengths \& Costs}}

\begin{table}[!ht] 
\renewcommand{\arraystretch}{1.15} 
\centering
\footnotesize
\captionsetup{font=footnotesize}
\vspace{-0.2cm}
\hspace{-0.2cm}
\begin{tabular}{@{\hskip 0.1cm}l@{\hskip 0.2cm}c@{\hskip 0.2cm}c@{\hskip 0.2cm}c@{\hskip 0.2cm}c}
\toprule
\textbf{Method} & \makecell{\textbf{Inp Token} \\ \textbf{Length ($\downarrow$)}} & \makecell{\textbf{Out Token} \\ \textbf{Length ($\downarrow$)}} & \makecell{\textbf{Total Cost} \\ \textbf{ in \$ ($\downarrow$)}} & \makecell{\textbf{EM} \\ \textbf{($\uparrow$)}} \\
\midrule
StructQA & \textbf{9012} & 3590 & 0.99 & 0.48 \\
\rowcolor{blue!10}
\name & 9388 & \textbf{2524} & \textbf{0.85} & \textbf{0.57} \\
\bottomrule
\end{tabular}
\vspace{-0.1cm}
\caption{\textbf{Performance and Cost Comparison:} Comparison of total input and output token lengths (across all LLM calls), total cost, and EM scores for different methods.}
\label{tab:performance_cost_comparison}
\vspace{-0.25cm}
\end{table}

\noindent \textcolor{\updatedText}{ In table \ref{tab:performance_cost_comparison} we report the average total input and output token lengths on HotpotQA dataset (w.r.t. 100 random samples from dev set). We also share the cost estimate w.r.t. the latest GPT-3.5 LLM (both as reader and triple extractor model)
}

\noindent \textcolor{\updatedText}{The average input and output token lengths when simply using the reader LLM are 1334 and 10, respectively. Its worth noting that for performance improvements beyond LLMs on complex NLP datasets such as HotpotQA and MuSiQue, structured information extraction is immensely useful, as can be seen from the improvements across metrics in Table \ref{tab:mhqa_results} of the paper.}

\subsection{Dataset Statistics}
\label{sec:appx_dataset_stats}
\begin{table}[!h]
  \centering
  \captionsetup{position=below} 
  \setlength{\abovecaptionskip}{10pt}
   
  \begin{tabular}{ >{\centering\arraybackslash} p{3cm} >{\centering\arraybackslash} p{1.5cm} >{\centering\arraybackslash} p{1.5cm}}
    \hline
    \textbf{Dataset} & \textbf{Train} & \textbf{Dev} \\
    \hline
    HotpotQA  & 90,447 &   7,405 \\
    MuSiQue & 19,938 & 2,417 \\
    \hline
  \end{tabular}
  \caption{\textbf{Train and Dev Statistics} - number of samples}
  \label{tab:QA_stats}
\end{table}

\begin{table}[!h]
  \centering

  \captionsetup{position=below} 
  \setlength{\abovecaptionskip}{10pt}
   
  \begin{tabular}{ >{\centering\arraybackslash} p{3.5cm} >{\centering\arraybackslash} p{3.5cm}}
    \hline
    \textbf{Dataset} & \textbf{number of instances} \\
    \hline
    MuSiQue 2-hop &  667 \\ 
    MuSiQue 3-hop &  366 \\ 
    MuSiQue 4-hop &  200 \\ 
    \hline
  \end{tabular}
  \caption{
Hop-wise distribution of questions used for evaluation in Section \ref{sec:results} for MuSiQue}
  \label{tab:QA_stats_hop-wise}
\end{table}

Table \ref{tab:QA_stats} provides details on the total number of training and development set samples in each dataset. From these, we have randomly sampled $10,000$ questions (unlabelled or unannotated) from the training set to create our auxiliary schema. For each dataset, we randomly sampled $1000$ data points from the development set for evaluation purposes (as illustrated in section \ref{sec:exp_dataset}).
Table \ref{tab:QA_stats_hop-wise} details the hop-wise question statistics of the MuSiQue dataset (HotpotQA dataset \cite{yang2018hotpotqa} does not provide hop-wise distribution of their dataset).

\subsection{Guidelines for Human Annotators}
\label{sec:appx_guidelines_human_annot}

In the evaluation phase of \name (Table \ref{tab:human_eval}), human annotators played a crucial role in assessing the accuracy of the predicted answers. The process was structured as follows:
\begin{itemize}[noitemsep]
    \item Annotators were presented with a set of inputs for each question, which included:
    \begin{itemize}[noitemsep]
        \item The question
        \item The gold answer
        \item The predicted answer generated by one of the MHQA methods
    \end{itemize}
    \item The primary task for the annotators was to determine the correctness of the predicted answer in comparison to the gold answer. This evaluation was binary, with annotators assigning:
    \begin{itemize}[noitemsep]
        \item A score of \textbf{0} if the predicted answer was deemed incorrect
        \item A score of \textbf{1} if it was considered correct
    \end{itemize}
    \item The evaluation process encompassed all methods: Base with supporting documents, StructQA, and \name, each evaluated with a total of 100 questions. Therefore:
    \begin{itemize}[noitemsep]
        \item Each annotator was responsible for evaluating a comprehensive total of \textbf{300 questions}, covering all possible combinations of questions and methods.
    \end{itemize}
\end{itemize}

\noindent The numbers in Table \ref{tab:human_eval} represent the average scores from three human annotators.

\subsection{Prompts used in \name}
\label{sec:appx_prompt}
In this section we detail the prompts used for solving different subtasks using LLMs.

\subsubsection{Entity Extraction from supporting documents}
\label{sec:prompt_enty_extr_supp_docs}
For extracting named entities from the supporting documents, we use the following prompt. Few-shot examples used to improve the performance of the LLM is also included in the prompt.

\begin{tcolorbox}[breakable, colframe=gray, colback=white, boxrule=0.5pt, sharp corners, width=\linewidth]

\texttt{\textbf{Task:} Extract ALL the named entities from the given sentence (extract all time intervals, names, dates, organizations and locations).}

       \vspace{0cm}    
           \texttt{\textbf{ Examples:} Use the following examples to understand the task better.}
               \vspace{0.3cm}    
            \texttt{\textbf{ Sentence:} William Rast is an American clothing line founded by Justin Timberlake and Trace Ayala.}
           
           \texttt{\textbf{  Entities:}
            William Rast, American, Justin Timberlake, Trace Ayala}
         
               \vspace{0.3cm}   
            \texttt{\textbf{ Sentence:} The Glennwanis Hotel is a historic hotel in Glennville, Georgia, Tattnall County, Georgia,
            built on the site of the Hughes Hotel.}
         
           \texttt{\textbf{  Entities:}
            Glennwanis Hotel, Glennville, Georgia, Tattnall County, Georgia, Hughes Hotel }

\end{tcolorbox}

\subsubsection{Entity and Relation Extraction from the Question}
\label{sec:prompt_enty_rel_extr_query}
Instead of separately extracting the entity and the relations from the inference query, we use a single LLM call. Below we detail the prompt along with the few-shot examples used.

\begin{tcolorbox}[breakable, colframe=gray, colback=white, boxrule=0.5pt, sharp corners,    width=\linewidth]

   \texttt{\textbf{Task:} extract all named entities from the above question. Then extract all the important information (each information should be 2-3 words) needed to answer the question.\
          Output format - ``entities: [ent1, ent2, ...]  important relations: [info1, info2, ...]". Please note, do not give named entities in the `important relations' }
 
        \texttt{\textbf{Use the following examples to understand the task:}}
        \vspace{0.3cm} \\
       \texttt{\textbf{ Question:} Who is the author of the book that inspired the movie starring Tom Hanks as a symbologist?} \\
        \texttt{\textbf{Entities:} [Tom Hanks]} \\
        \texttt{\textbf{Important Relations:} [author of,  inspired the movie, stars,as a symbologist] }
        
         \vspace{0.2cm}
        \texttt{\textbf{ Question:} Did the company that Elon Musk co-founded in 2002 eventually merge with a firm that had been contracted by NASA to resupply the International Space Station? } \\\
        \texttt{\textbf{Entities:} [`Elon Musk', `2002', `NASA', `International Space Station'] }
       \texttt{\textbf{ Important Relations:} [`co-founded', `merge with', `contracted by', `resupply'] }
        
\end{tcolorbox}

\subsubsection{Subject and Object Entity type Estimation}
\label{sec:prompt_enty_type_est}
As discussed in Section \ref{subsec: knowledge schema construction} and shown in Figure \ref{fig:approach}, to create the knowledge schema from the query, we need to estimate the subject and object entity types corresponding to a given relation. We use an LLM for the same and report the prompt below.

\begin{tcolorbox}[breakable, colframe=gray, colback=white, boxrule=0.5pt, sharp corners,   after={\vskip-\lastskip\hrule height 0pt},  width=\linewidth]

\texttt{\textbf{Task:} Generate the subject and object entity type for the following relation, w.r.t. the provided context information. } \\
\texttt{\textbf{relation: }{input}}
\\ 
\texttt{\textbf{Context:} {context} ; DO NOT extract partial info.}
\\
     \texttt{\textbf{Output format:} [<subject entity type>, <object entity type>] }

\end{tcolorbox}

\subsubsection{Triples Extraction from supporting documents}
\label{sec:prompt_triples_extr_docs}
Here we report the prompt along with the few-shot examples we used for knowledge triple extraction from raw text.

\begin{tcolorbox}[breakable, colframe=gray, colback=white, boxrule=0.5pt, sharp corners,   after={\vskip-\lastskip\hrule height 0pt},  width=\linewidth]

\texttt{\textbf{Task:} Comprehensively extract ALL the triples (subject, relation, object) from below given paragraph. Ensure that the subject and objects in the triples are named entities (name of person, organization, dates etc) and not multiple in number. You will be HEAVILY PENALIZED if you violate this constraint.}
 
           \texttt{\textbf{ Examples:} Use the following examples to understand the task better. } \
            \texttt{\textbf{Paragraph:} William Rast is an American clothing line founded by Justin Timberlake and Trace Ayala.
            It is most known for their premium jeans.  On October 17, 2006, Justin Timberlake and Trace Ayala put on their
            first fashion show to launch their new William Rast clothing line.  The label also produces other clothing items
            such as jackets and tops.  The company started first as a denim line, later evolving into a men's and women's clothing line.}
            
           \texttt{\textbf{ Triples:}
            \begin{enumerate}[label=\roman*., itemsep=0pt, parsep=0pt, partopsep=0pt]
            \item subject: William Rast, relation: clothing line, object: American
            \item subject: William Rast, relation: founded by, object: Justin Timberlake
            \item subject: William Rast, relation: founded by, object: Trace Ayala
            \item subject: William Rast, relation: known for, object: premium jeans
            \item subject: William Rast, relation: launched on , object: October 17, 2006
            \item subject: Justin Timberlake, relation: first fashion show, object: October 17, 2006
            \item subject: Trace Ayala, relation: first fashion show, object: October 17, 2006
            \item subject: William Rast, relation: produces, object: jackets
           \item subject: William Rast, relation: produces, object: tops
            \item subject: William Rast, relation: started as, object: denim line
            \item subject: William Rast, relation: evolved into, object: men's and women's clothing line
            \end{enumerate}
            }
 
            \texttt{\textbf{ Paragraph:} The Glennwanis Hotel is a historic hotel in Glennville, Georgia, Tattnall County, Georgia, built on the site of the Hughes Hotel.  The hotel is located at 209-215 East Barnard Street.  The old Hughes Hotel was
            built out of Georgia pine circa 1905 and burned in 1920.  The Glennwanis was built in brick in 1926.  The local Kiwanis
            club led the effort to get the replacement hotel built, and organized a Glennville Hotel Company with directors being
            local business leaders.  The wife of a local doctor won a naming contest with the name ``Glennwanis Hotel", a suggestion
            combining ``Glennville" and "Kiwanis}
 
                   \texttt{\textbf{ Triples:}
            \begin{enumerate}[label=\roman*., itemsep=0pt, parsep=0pt, partopsep=0pt]
            \item subject: Glennwanis Hotel, relation: is located in, object: 209-215 East Barnard Street, Glennville, Tattnall County, Georgia
            \item subject: Glennwanis Hotel, relation: was built on the site of, object: Hughes Hotel
            \item subject: Hughes Hotel, relation: was built out of, object: Georgia pine
            \item subject: Hughes Hotel, relation: was built circa, object: 1905
            \item subject: Hughes Hotel, relation: burned in, object: 1920
            \item subject: Glennwanis Hotel, relation: was re-built in, object: 1926
            \item subject: Glennwanis Hotel, relation: was re-built using, object: brick
            \item subject: Kiwanis club, relation: led the effort to re-build, object: Glennwanis Hotel
            \item subject: Kiwanis club, relation: organized, object: Glennville Hotel Company
            \item subject: Glennville Hotel Company, relation: directors, object: local business leaders
            \item subject: Glennwanis Hotel, relation: combines, object: ``Glennville" and ``Kiwanis"
            \end{enumerate}
            }

\end{tcolorbox}

\begin{figure*}
  \centering
  \includegraphics[width=1.0\textwidth]{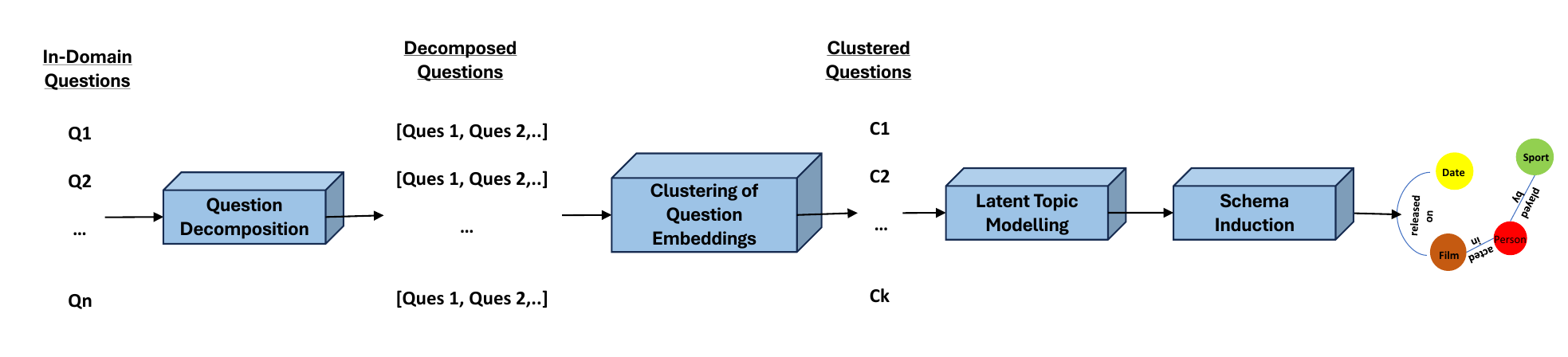}
  \caption{\textbf{Auxilliary Graph Schema Construction Overview } }
  \label{fig:graph-schema}
\end{figure*}

\subsubsection{LLM Reader}
\label{sec:prompt_reader_LLM}

Below we report the reader LLM prompt. Please note that we instruct the LLM to reason in three steps - (i) relevant fact extraction (ii) reasoning based on the relevant facts (iii) final response (with an option to request details about named entities) 

\begin{tcolorbox}[breakable, colframe=gray, colback=white, boxrule=0.5pt, sharp corners,   after={\vskip-\lastskip\hrule height 0pt},  width=\linewidth]

  \texttt{\textbf{Question:\{question\}} \\
         Read the above question carefully, understand the answer category.
         Now, given you understand the question, use the following facts to answer the question. Please note that the meaning  of facts is HEAVILY dependent on the context or the document from which it was extracted. 
         {prunned\_and\_verbalized\_hyperkg}
         (Note: the above set of facts can be noisy, so if ambiguous information is present then focus on the question and keywords in the question - {relations\_in\_question})
         First, fetch the relevant set of facts from the above taking into consideration the context of the fact (DO NOT generate your facts),\
         then by combining them answer the question - {question}. } \\
           \texttt{\textbf{ Instruction:} Give the answer in the following format - } \\
       \texttt{\textbf{  Relevant facts:} <facts (and context of those facts) relevant to the question> } \\
         \texttt{\textbf{  Reasoning:} <Here, give your thought process of how you use the named entities - {named\_entities\_in\_question}, relations - {relations\_in\_question} of the question and accordingly traverse the facts given above to arrive at your answer> } \\
          \texttt{\textbf{ Final Response:} <If all the necessary facts are available to answer the question, express it in 3-4 words - Else, say None> } \\
           \texttt{\textbf{Further query:} <if `Final Response' is None, then think step-by-step using the `Reasoning' and other facts and state which named entity's direct information is missing>}

\end{tcolorbox}


 \subsection{Auxilliary Schema Construction}
\label{sec:auxiliary_graph_schema_creation}
As stated in Section \ref{subsec:schema_creation}, we use a collection of in-domain questions to construct a graph schema (See definition in Section \ref{sec: prelim}), that captures the global level blueprint of information required to answer questions in the target domain. In this section, we detail the procedure used to create the same. We employ a four-step process, starting from the un-annotated in-domain multi-hop questions.

\subsubsection{Question Decomposition}
Given the complexity of multi-hop questions and the challenges in processing their embeddings, we first decompose these into simpler, single-hop questions. This decomposition is facilitated by the \texttt{llama-13b} LLM, which effectively breaks down complex questions into their constituent single-hop components. Decomposing $10,000$ questions using SoTA LLMs is very expensive, thus we opted for a smaller, open-weight LLM. We use few-shot prompting to ensure the model understands the task well. We use the following prompt for the same:

\begin{tcolorbox}[breakable, colframe=gray, colback=white, boxrule=0.5pt, sharp corners,   after={\vskip-\lastskip\hrule height 0pt},  width=\linewidth]
\noindent \texttt{\textbf{Task:} Decompose a multi-hop question into a series of single-hop questions to assist in finding the answer in a step-by-step manner.}

\vspace{0cm}
\noindent \texttt{\textbf{Instruction:} 
\begin{enumerate}[label=\roman*., itemsep=0pt, parsep=0pt, partopsep=0pt]
\item Proceed with the decomposition according to the outlined chain of thought. 
\item Only return the sub-questions, nothing else. 
\item Focus on the entities in the question 
\item If decomposition is not clear, do not guess.
\end{enumerate}
}

\vspace{0cm}
\noindent \texttt{
\textbf{Question:} Lily's Driftwood Bay premiered on what British television channel that is operated by a joint venture between Viacom International Media Networks Europe and Sky plc?}

    \vspace{0cm}
\noindent \texttt{\textbf{Chain of Thought Instructions}:
\begin{enumerate}[itemsep=0pt, parsep=0pt, partopsep=0pt]
 \item Identify the ultimate goal of the question: Find the television channel on which Lily's Driftwood Bay premiered.
\item Recognize that there are two pieces of information needed:
  a) Determine the British television channel operated by the specific joint venture.
  b) Find out if Lily's Driftwood Bay premiered on this channel.
\item Create single-hop sub-questions that will answer these pieces of information separately.
\end{enumerate}}

\noindent \texttt{
\textbf{Output}:
\begin{enumerate}[itemsep=0pt, parsep=0pt, partopsep=0pt]
\item Which British television channel is operated by a joint venture between Viacom International Media Networks Europe and Sky plc?
\item Did Lily's Driftwood Bay premiere on this specific British television channel?
\end{enumerate}
}
\end{tcolorbox}

\subsubsection{Cluster Question Embeddings}
We then apply K-Means clustering to the embeddings of the decomposed single-hop questions to group similar questions. We use OpenAI embedding model (\texttt{text-embedding-ada-002}) to compute text embeddings. The optimal number of clusters (\(k\)) is determined by evaluating the clustering quality through a combined clustering evaluation score (CCES) which is a weighted sum of three normalized scores (normalized to lie between 0 and 1): silhouette score (\(S_{\text{norm}}\)), inverted Davies-Bouldin score (\(DB_{\text{norm}}\)), and Calinski-Harabasz score (\(CH_{\text{norm}}\)), with equal weightage assigned to each. These scores collectively assess the cohesion, separation, compactness, distinctness, and dispersion of the clusters. We use the elbow method for determining the optimal number of clusters. We report the elbow plot below.

\begin{figure}[!ht]
  \centering
\includegraphics[width=0.8\columnwidth]{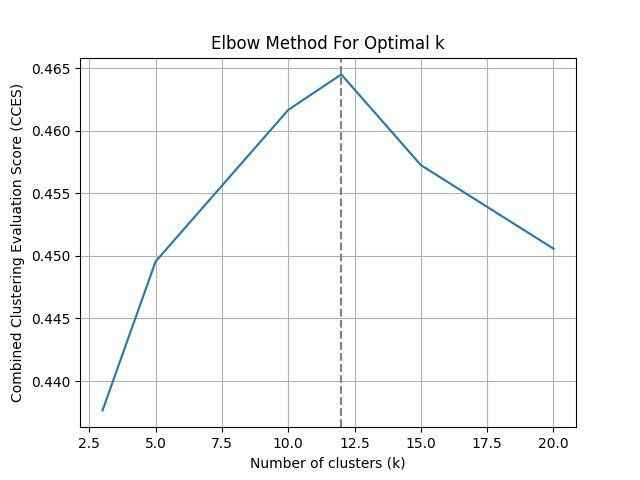}
  \caption{\textbf{Elbow Plot} for Kmeans clustering on $10000$ questions from HotpotQA dataset.}
  \label{fig:clustering_elbow_plot}
\end{figure}

\noindent Below, we report a couple of cluster centroids with a few data points (questions) around the centroid.

\textbf{Cluster 1}
\begin{itemize}[noitemsep]
    \item \textit{Which football team did Kevin John Ufuoma Akpoguma play for?}
    \item \textit{Who is the former Houston Texans head coach?}
    \item \textit{Which NFL team currently employs him?}
\end{itemize}

\textbf{Cluster 2}
\begin{itemize}[noitemsep]
    \item \textit{What is the name of the actress?}
    \item \textit{Who are Rohan Bopanna and Cara Black?}
    \item \textit{What is the name of the protagonist in the show "Pretty Little Liars"?}
\end{itemize}
As can be seen from the above clusters, they capture latent topics in the dataset such as sports (cluster 1) and movies (cluster 2).

\subsubsection{Latent Question Category Modelling}
The next step involves latent topic modeling to identify question categories within the single-hop question clusters. By extracting the cluster centroid question and the five questions closest to each centroid, we prompt an LLM (\texttt{gpt-4-1106-preview} in this case) to discern latent topics, focusing on entity types and relations rather than specific entities. The prompt we use deducing the latent question category is as follows:

\begin{tcolorbox}[colframe=gray, colback=white, boxrule=0.5pt, sharp corners, width=\linewidth]
\noindent \texttt{\textbf{Task:} Extract latent topics with focus on entity types (not the entities themselves) and relations in below given questions. These topics should be the underlying question categories for these questions.}

\vspace{0cm}
\noindent \texttt{\textbf{Instructions:}
\begin{enumerate}[label=\roman*., itemsep=0pt, parsep=0pt, partopsep=0pt]
\item Keep in mind that I will later use these latent topics to create a KG schema such that the KG can answer questions such as the ones listed below.
\item Answer in 1 sentence using detailed and informative words only. 
\end{enumerate}
}
\end{tcolorbox}

\begin{tcolorbox}[colframe=gray, colback=white, boxrule=0.5pt, sharp corners, width=\linewidth]
\texttt{
\begin{enumerate}[label=\roman*., itemsep=0pt, parsep=0pt, partopsep=0pt]
\setcounter{enumi}{2}
\item Consider cross-question information too for determining the latent topics.
\item Directly give the latent topics without preamble text like ``the latent topic is ...", and use bullets to separate topics.
\end{enumerate}
}
\end{tcolorbox}

\subsubsection{Schema Induction}
Finally, we prompt the LLM to generate a graph schema based on the broad question categories identified in the previous step. This schema is explicitly designed as an information organization blueprint relevant to the identified question categories, thereby facilitating a structured approach to answering questions within the domain. We use the following prompt for the same - 

\begin{tcolorbox}[colframe=gray, colback=white, boxrule=0.5pt, sharp corners, width=\linewidth]

\noindent \texttt{\textbf{Task:} Create a graph schema for my KG using the broad question categories that I want to be answered by the KG.} 

\noindent \texttt{The broad question categories are:
 \{question categories\}}

\vspace{0cm}
\noindent \texttt{\textbf{Instructions:}
\begin{enumerate}[label=\roman*., itemsep=0pt, parsep=0pt, partopsep=0pt]
\item Create a set of relations for a knowledge graph that clearly and unambiguously express the relationships between entities ensuring they are reusable, standardized, and semantically meaningful.
\item Generate a list of distinct, relevant, and comprehensive entities for a knowledge graph about wikipedia text, ensuring they are specific, meaningful, and cover all aspects of the domain.
\item  After extracting the entities and relation types, for the graph schema of the knowledge graph return the all the triplets ``entity type - Relation Type - Entity Type" in the output as a list
\end{enumerate}
}
\end{tcolorbox}

\begin{tcolorbox}[colframe=gray, colback=white, boxrule=0.5pt, sharp corners, width=\linewidth]
\texttt{
\begin{enumerate}[label=\roman*., itemsep=0pt, parsep=0pt, partopsep=0pt]
\setcounter{enumi}{3}
\item  Important: ensure that the triplets generated forms a connected graph
\item Important: - nothing else other than the list of triplets should be returned. Format for the list: 
[(``Entity type", ``relation", ``Entity type"),
(``Entity type", ``relation", ``Entity type"),...]
\end{enumerate}
}
\end{tcolorbox}

\subsection{Case Study of \name on HotpotQA dataset}
\label{sec:appx_case_study}

In this section, we showcase our constructed hyper-relational KG extracted from the supporting documents. We illustrate the corresponding reasoning process, highlighting relevant facts, and ultimately derive the answer for the question.
\\
Now, consider the following example question from the HotpotQA dataset.

\vspace{0.2cm}
\noindent \textbf{Question:} \textit{`What major truck road is located in Backford Cross?'}
\vspace{0.2cm}



\noindent Based on the question, our method discovers related supporting documents and then fetches structured information from them (Section \ref{subsec:HyperKGConstruction}). Below we list those discovered supporting documents.

\vspace{0.2cm}

\noindent \textbf{Supporting Document: 1}

\noindent \textbf{Title:} \textit{Backford Cross}

\noindent \textbf{Passage:} \textit{Backford Cross is a village on the Wirral Peninsula, Cheshire, England.  It is a suburb of the town of Ellesmere Port and part of Cheshire West and Chester.  Backford Cross is located around the A41/A5117 junction, south of Great Sutton and about 1.5 mi north of the village of Backford, near Chester.  Backford Cross is largely made up of residential homes built from 1990 onwards and serves as a commuter village to Ellesmere Port and Chester, although inhabitants show no allegiance to either locality.  The area is split between postcode districts, with parts of the village in Great Sutton, Ellesmere Port CH66 and other areas in Backford, Chester CH1.}
\vspace{0.3cm}

\noindent \textbf{Supporting Document: 2}

\noindent \textbf{Title:} \textit{A5117 road}

\noindent \textbf{Passage:}  \textit{The A5117 is a road in Cheshire, England.  It runs between Shotwick ( ) and Helsby ( ) and connects the A550 at Woodbank to the M56.  As such it forms a northerly bypass to Chester and a shorter route between the North West and North Wales than the A55.  The road is dualled west of the M56.  There is roundabout with the A540 and at Dunkirk at the western terminus of the M56.  East of the junction the road is single carriageway and crosses the A41 by way of a roundabout at Backford Cross.  The A5117 intersects the M53 at Junction 10.  This junction is just east of Cheshire Oaks Designer Outlet.  The road then continues almost parallel to the M56, which it intersects at Junction 14, at which there is a Motorway service area.  The road then continues south east to terminate where it joins the A56 at Helsby.}
\vspace{0.3cm}

\noindent \textbf{Supporting Document: 3}

\noindent \textbf{Title:} \textit{Strawberry Park, Cheshire}

\noindent \textbf{Passage:}   \textit{Strawberry Park and Strawberry Fields are suburbs in the town of Ellesmere Port, Cheshire West and Chester.  They are located to the south of Hope Farm and to the west of Backford Cross.}
\vspace{0.3cm}

\noindent \textbf{Supporting Document: 4}

\noindent \textbf{Title:} \textit{A41 road}

\noindent \textbf{Passage:}   \textit{The A41 is a major trunk road in England that links London and Birkenhead, although it has now in parts been superseded by motorways.  It passes through or near various towns and cities including Watford, Kings Langley, Hemel Hempstead, Aylesbury, Solihull, Birmingham, West Bromwich, Wolverhampton, Newport, Whitchurch, Chester and Ellesmere Port.}
\vspace{0.3cm}


\noindent From these supporting documents, we create the Hyper-relational KG and further refine this graph to retain query relevant facts (Section \ref{subsec: knowledge schema construction}). Below we report the same (\textcolor{\updatedText}{blue} colored text refers to the additional attributes of the hyper triple, \textcolor{red}{red} colored text refers to the subject entity, \textcolor{green}{green} colored text refers to the relation, and \textcolor{purple}{purple} colored text refers to the object entity). 

\vspace{0.2cm}


\textbf{Distilled Hyper-Relational KG:}

\begin{itemize}
    \item \textit{\textcolor{\updatedText}{\textit{context}}: `Strawberry Park, Cheshire' \\
    \textcolor{red}{\textit{subject}}: `Strawberry Park' \\
    \textcolor{green!60!black}{\textit{relation}}: `is a suburb in' \\
    \textcolor{purple}{\textit{object}}: `Ellesmere Port, Cheshire West'}
    
    \item \textit{\textcolor{\updatedText}{\textit{context}}: `Strawberry Park, Cheshire' \\
    \textcolor{red}{\textit{subject}}: `Strawberry Fields' \\
    \textcolor{green!60!black}{\textit{relation}}: `is a suburb in' \\
    \textcolor{purple}{\textit{object}}: `Ellesmere Port, Cheshire West'}
    
    \item \textit{\textcolor{\updatedText}{\textit{context}}: `Strawberry Park, Cheshire' \\
    \textcolor{red}{\textit{subject}}: `Ellesmere Port' \\
    \textcolor{green!60!black}{\textit{relation}}: `located in' \\
    \textcolor{purple}{\textit{object}}: `Cheshire West and Chester'}
    
    \item \textit{\textcolor{\updatedText}{\textit{context}}: `Strawberry Park, Cheshire' \\
    \textcolor{red}{\textit{subject}}: `Strawberry Park' \\
    \textcolor{green!60!black}{\textit{relation}}: `located to the south of' \\
    \textcolor{purple}{\textit{object}}: `Hope Farm'}
    
    \item \textit{\textcolor{\updatedText}{\textit{context}}: `Backford Cross' \\
    \textcolor{red}{\textit{subject}}: `Backford Cross' \\
    \textcolor{green!60!black}{\textit{relation}}: `is located in' \\
    \textcolor{purple}{\textit{object}}: `Wirral Peninsula, Cheshire, England'}
    
    
    
    
    
    
    \item \textit{\textcolor{\updatedText}{\textit{context}}: `A5117 road' \\
    \textcolor{red}{\textit{subject}}: `A5117' \\
    \textcolor{green!60!black}{\textit{relation}}: `connects' \\
    \textcolor{purple}{\textit{object}}: `A550 at Woodbank to M56'}
    
    \item \textit{\textcolor{\updatedText}{\textit{context}}: `A5117 road' \\
    \textcolor{red}{\textit{subject}}: `A5117' \\
    \textcolor{green!60!black}{\textit{relation}}: `forms a bypass to' \\
    \textcolor{purple}{\textit{object}}: `Chester'}
    
    \item \textit{\textcolor{\updatedText}{\textit{context}}: `A5117 road' \\
    \textcolor{red}{\textit{subject}}: `A5117' \\
    \textcolor{green!60!black}{\textit{relation}}: `forms a shorter route between' \\
    \textcolor{purple}{\textit{object}}: `North West and North Wales'}
    
    \item \textit{\textcolor{\updatedText}{\textit{context}}: `A5117 road' \\
    \textcolor{red}{\textit{subject}}: `A5117' \\
    \textcolor{green!60!black}{\textit{relation}}: `is dualled west of' \\
    \textcolor{purple}{\textit{object}}: `M56'}
    
    \item \textit{\textcolor{\updatedText}{\textit{context}}: `A5117 road' \\
    \textcolor{red}{\textit{subject}}: `A5117' \\
    \textcolor{green!60!black}{\textit{relation}}: `has a roundabout with' \\
    \textcolor{purple}{\textit{object}}: `A540'}
    
    \item \textit{\textcolor{\updatedText}{\textit{context}}: `A5117 road' \\
    \textcolor{red}{\textit{subject}}: `A5117' \\
    \textcolor{green!60!black}{\textit{relation}}: `has a roundabout at' \\
    \textcolor{purple}{\textit{object}}: `Dunkirk'}
    
    \item \textit{\textcolor{\updatedText}{\textit{context}}: `A5117 road' \\
    \textcolor{red}{\textit{subject}}: `A5117' \\
    \textcolor{green!60!black}{\textit{relation}}: `intersects' \\
    \textcolor{purple}{\textit{object}}: `M53 at Junction 10'}
    
    \item \textit{\textcolor{\updatedText}{\textit{context}}: `A5117 road' \\
    \textcolor{red}{\textit{subject}}: `M53' \\
    \textcolor{green!60!black}{\textit{relation}}: `is intersected by' \\
    \textcolor{purple}{\textit{object}}: `A5117 at Junction 10'}
    
    \item \textit{\textcolor{\updatedText}{\textit{context}}: `A41 road' \\
    \textcolor{red}{\textit{subject}}: `A41' \\
    \textcolor{green!60!black}{\textit{relation}}: `is a major trunk road in' \\
    \textcolor{purple}{\textit{object}}: `England'}
    
    \item \textit{\textcolor{\updatedText}{\textit{context}}: `A41 road' \\
    \textcolor{red}{\textit{subject}}: `A41' \\
    \textcolor{green!60!black}{\textit{relation}}: `links' \\
    \textcolor{purple}{\textit{object}}: `London and Birkenhead'}
    
    \item \textit{\textcolor{\updatedText}{\textit{context}}: `A41 road' \\
    \textcolor{red}{\textit{subject}}: `A41' \\
    \textcolor{green!60!black}{\textit{relation}}: `superseded by' \\
    \textcolor{purple}{\textit{object}}: `motorways'}
    
    \item \textit{\textcolor{\updatedText}{\textit{context}}: `A41 road' \\
    \textcolor{red}{\textit{subject}}: `A41' \\
    \textcolor{green!60!black}{\textit{relation}}: `passes through' \\
    \textcolor{purple}{\textit{object}}: `Watford'}
    
    \item \textit{\textcolor{\updatedText}{\textit{context}}: `A41 road' \\
    \textcolor{red}{\textit{subject}}: `A41' \\
    \textcolor{green!60!black}{\textit{relation}}: `passes near' \\
    \textcolor{purple}{\textit{object}}: `Kings Langley'}
    
    \item \textit{\textcolor{\updatedText}{\textit{context}}: `A41 road' \\
    \textcolor{red}{\textit{subject}}: `A41' \\
    \textcolor{green!60!black}{\textit{relation}}: `passes near' \\
    \textcolor{purple}{\textit{object}}: `Hemel Hempstead'}
    
    \item \textit{\textcolor{\updatedText}{\textit{context}}: `A41 road' \\
    \textcolor{red}{\textit{subject}}: `A41' \\
    \textcolor{green!60!black}{\textit{relation}}: `passes near' \\
    \textcolor{purple}{\textit{object}}: `Aylesbury'}
    
    \item \textit{\textcolor{\updatedText}{\textit{context}}: `A41 road' \\
    \textcolor{red}{\textit{subject}}: `A41' \\
    \textcolor{green!60!black}{\textit{relation}}: `passes near' \\
    \textcolor{purple}{\textit{object}}: `Solihull'}
    
    \item \textit{\textcolor{\updatedText}{\textit{context}}: `A41 road' \\
    \textcolor{red}{\textit{subject}}: `A41' \\
    \textcolor{green!60!black}{\textit{relation}}: `passes through' \\
    \textcolor{purple}{\textit{object}}: `Birmingham'}
    
    \item \textit{\textcolor{\updatedText}{\textit{context}}: `A41 road' \\
    \textcolor{red}{\textit{subject}}: `A41' \\
    \textcolor{green!60!black}{\textit{relation}}: `passes near' \\
    \textcolor{purple}{\textit{object}}: `West Bromwich'}
    
    \item \textit{\textcolor{\updatedText}{\textit{context}}: `A41 road' \\
    \textcolor{red}{\textit{subject}}: `A41' \\
    \textcolor{green!60!black}{\textit{relation}}: `passes near' \\
    \textcolor{purple}{\textit{object}}: `Wolverhampton'}
    
    \item \textit{\textcolor{\updatedText}{\textit{context}}: `A41 road' \\
    \textcolor{red}{\textit{subject}}: `A41' \\
    \textcolor{green!60!black}{\textit{relation}}: `passes through' \\
    \textcolor{purple}{\textit{object}}: `Newport'}
\end{itemize}

\noindent This distilled graph (after verbalization and proper formatting 
 (Section \ref{sec:input_prompt})) is provided as input, along with the question, to the reader LLM. For this example, we use the SoTA, \texttt{gpt-4-1106-preview} as the reader LLM. Below, we describe the LLMs response which comprises of three things: 
 \begin{itemize}[noitemsep]
     \item \textbf{Relevant facts} set of hyper triples which the LLM thought were important for the answering the question. These are verbalized hyper triples as the input to the LLM was also verbalized hyper triples.
     \item \textbf{Reasoning}: LLMs thought process of going from the distilled hyper-relational KG to the final answer
     \item \textbf{Final answer}: answer predicted by the LLM.
 \end{itemize}
\vspace{0.2cm}

\noindent \textbf{Our Method's Response}

\vspace{0.1cm}

\noindent \textit{Relevant facts:}
    
\textit{- \textcolor{red}{Backford Cross} \textcolor{green!60!black}{is located around} \textcolor{purple}{A41/A5117 junction} (This fact was extracted from document named \textcolor{\updatedText}{"Backford Cross "})}

\textit{- \textcolor{red}{A41} \textcolor{green!60!black}{is a major trunk road in} \textcolor{purple}{England} (This fact was extracted from document named \textcolor{\updatedText}{"A41 road "})}

\textit{- \textcolor{red}{A5117} \textcolor{green!60!black}{is a road in} \textcolor{purple}{Cheshire, England} (This fact was extracted from document named \textcolor{\updatedText}{"A5117 road "})}

\vspace{0.1cm}
\noindent \textit{Reasoning:}
\vspace{0.1cm}

\textit{The question asks for a major truck road located in Backford Cross. The fact that Backford Cross is located around the A41/A5117 junction indicates that these are the major roads in that area. The A41 is specifically described as a major trunk road, which fits the description of a ``major truck road." The A5117 is also mentioned as a road in Cheshire, but it is not specified as a major trunk road. Therefore, based on the information provided, the A41 is the major truck road located in Backford Cross.}

\vspace{0.2cm}

\noindent \textit{Final Answer:}

\vspace{0.1cm}

\textit{A41}


\end{document}